\documentclass[dvipsnames]{article} %
\usepackage{colm2024_conference}
\pdfoutput=1
\setcitestyle{numbers,square,citesep={,}}

\usepackage{booktabs}
\usepackage{graphicx}
\usepackage{enumitem}
\usepackage{wrapfig}
\usepackage{algorithm}
\usepackage{algpseudocode}
\usepackage{microtype}
\usepackage{amsmath}
\usepackage{amsthm}
\usepackage{colortbl}
\usepackage[utf8]{inputenc}
\definecolor{lightgray}{rgb}{0.9,0.9,0.9}
\usepackage{caption}
\usepackage{subcaption}
\usepackage{setspace}
\usepackage{url}
\usepackage{multirow}
\usepackage{colortbl}
\usepackage{tabularx}
\usepackage{blindtext}
\usepackage{pgfplots}
\pgfplotsset{compat=1.18} 
\usepackage{tikz}
\usetikzlibrary{er,positioning,bayesnet}
\usepackage{makecell}
\usepackage{tipa}
\usepackage{siunitx}
\usepackage{nicefrac}
\usepackage{tocloft}
\usepackage{listings}
\usepackage{cleveref}
\usepackage{placeins}
\usepackage{rotating}
\usepackage{silence}

\usepackage{tablefootnote}
\usepackage{threeparttable}
\usepackage[most]{tcolorbox}

\usepackage{xltabular}
\usepackage{adjustbox}
\usepackage{xurl}
\usepackage[normalem]{ulem}
\useunder{\uline}{\ul}{}
\usepackage{newtxtext,newtxmath}
\usepackage{CJKutf8}
\usepackage{pifont}
\usepackage{xcolor}

\definecolor{darkgreen}{rgb}{0.0, 0.6, 0.0}

\definecolor{pyellow}{rgb}{1.0, 1.0, 0.8} 

\usepackage[table]{xcolor}
\definecolor{pyellow}{RGB}{255,245,200}

\usepackage{amsmath,amsfonts,bm}

\def\eqref#1{equation~\ref{#1}}

\def\1{\bm{1}}

\DeclareMathAlphabet{\mathsfit}{\encodingdefault}{\sfdefault}{m}{sl}
\SetMathAlphabet{\mathsfit}{bold}{\encodingdefault}{\sfdefault}{bx}{n}

\newcommand*\justify{%
  \fontdimen2\font=0.4em%
  \fontdimen3\font=0.2em%
  \fontdimen4\font=0.1em%
  \fontdimen7\font=0.1em%
  \hyphenchar\font=`\-%
}

\renewcommand{\texttt}[1]{%
  \begingroup
  \ttfamily
  \begingroup\lccode`~=`/\lowercase{\endgroup\def~}{/\discretionary{}{}{}}%
  \begingroup\lccode`~=`[\lowercase{\endgroup\def~}{[\discretionary{}{}{}}%
  \begingroup\lccode`~=`.\lowercase{\endgroup\def~}{.\discretionary{}{}{}}%
  \catcode`/=\active\catcode`[=\active\catcode`.=\active
  \justify\scantokens{#1\noexpand}%
  \endgroup
}

\title{OmegaUse-OfficeVal: Benchmarking LLM Agents on Long-Horizon Office-Suite Tasks with Economic Grounding}

\author{
 \centering
 \small{}
 \textbf{Jingbo Zhou$^{*\dagger}$ \hspace{4mm} Yusai Zhao$^{*\dagger}$ \hspace{4mm}
 Qi Bao \hspace{4mm} Jingjia Cao \hspace{4mm} Zhenghai Chen \hspace{4mm} Chang Gao \\
 Kaiqi Guo \hspace{4mm} Muxin Guo \hspace{4mm} Mingxuan Li \hspace{4mm} Xinjiang Lu \hspace{4mm} Yanru Ma \hspace{4mm} Yixiong Xiao \\
 Zenghui Zhang \hspace{4mm} Le Zhang \hspace{4mm} Hua Wu$^{\dagger}$} \\
\vspace{1em}
 \centering
 \small{}
  Agent Frontier Team, Large Model Frontier Research Department, Baidu Inc., China. \\
  \vspace{0.45em}
  $^{*}$~Equal contribution, \quad $^{\dagger}$~Project co-lead and corresponding authors. \\[0.45em]
  Correspondence to: \texttt{\{zhoujingbo, zhaoyusai, wu\_hua\}@baidu.com}
}

\makeatletter\def\@abstract{
Large language model (LLM) agents are increasingly expected to assist users in completing tasks. However, existing benchmarks provide limited support for evaluating whether agents can carry out office-suite workflows at a reasonable cost.
We introduce {OmegaUse-OfficeVal}, a benchmark for evaluating LLM agents on long-horizon office-suite tasks with task-level economic grounding. The benchmark comprises 100 tasks derived from office-suite requests proposed by practitioners and adapted through a privacy-preserving process. On average, these tasks require 2.32 hours of human labor to complete.
An important feature of the benchmark is that each task is paired with two economic signals: human labor time and task price proxy. These signals enable direct comparisons between human costs and LLM inference costs, as well as value-weighted evaluation. To support stable evaluation, we develop code-based verifiers from fine-grained rubrics.
We evaluate several frontier LLMs together with a human baseline. Although all evaluated LLMs are substantially cheaper and faster than human workers, they have not yet approached human-level deliverable quality. The code and dataset are fully open-sourced, and more information is available on our project website: \url{https://omegause-officeval.github.io}. 
}\makeatother

\begin{document}
\begin{CJK}{UTF8}{gkai}
\maketitle
% 抵消模板 tcolorbox 的部分 after skip
\vspace{-0.2cm}

\begingroup
\captionsetup{
    type=figure,
    font=small,
    skip=3pt
}
\captionsetup[subfigure]{
    font=footnotesize,
    justification=centering,
    singlelinecheck=false,
    skip=3pt
}

\noindent
\begin{minipage}{\textwidth}
\centering

\begin{subfigure}[c]{0.38\linewidth}
    \centering
    \includegraphics[
        width=\linewidth,
        keepaspectratio
    ]{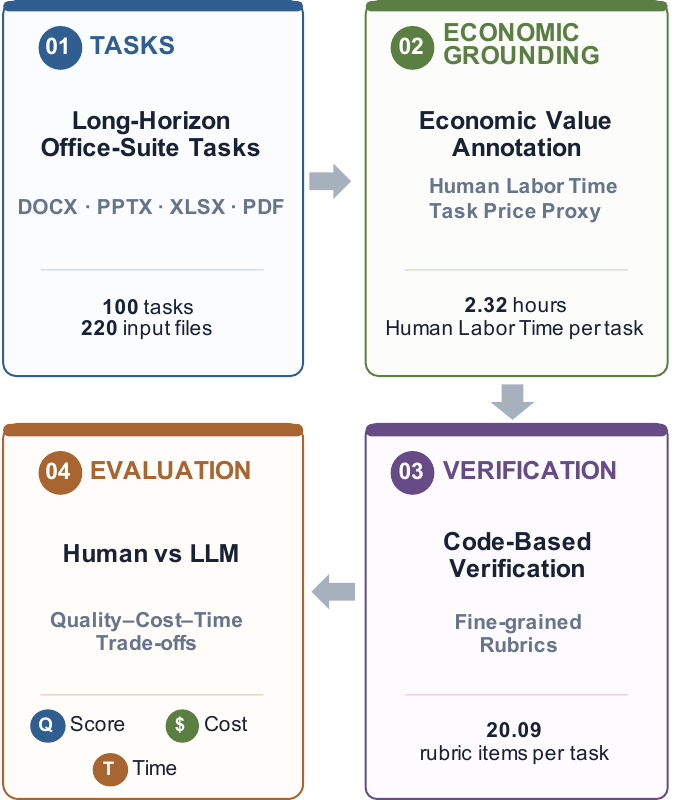}
    \caption{}
    \label{fig:homepage_pipeline}
\end{subfigure}
\hfill
\begin{subfigure}[c]{0.61\linewidth}
    \centering
    \includegraphics[
        width=\linewidth,
        keepaspectratio
    ]{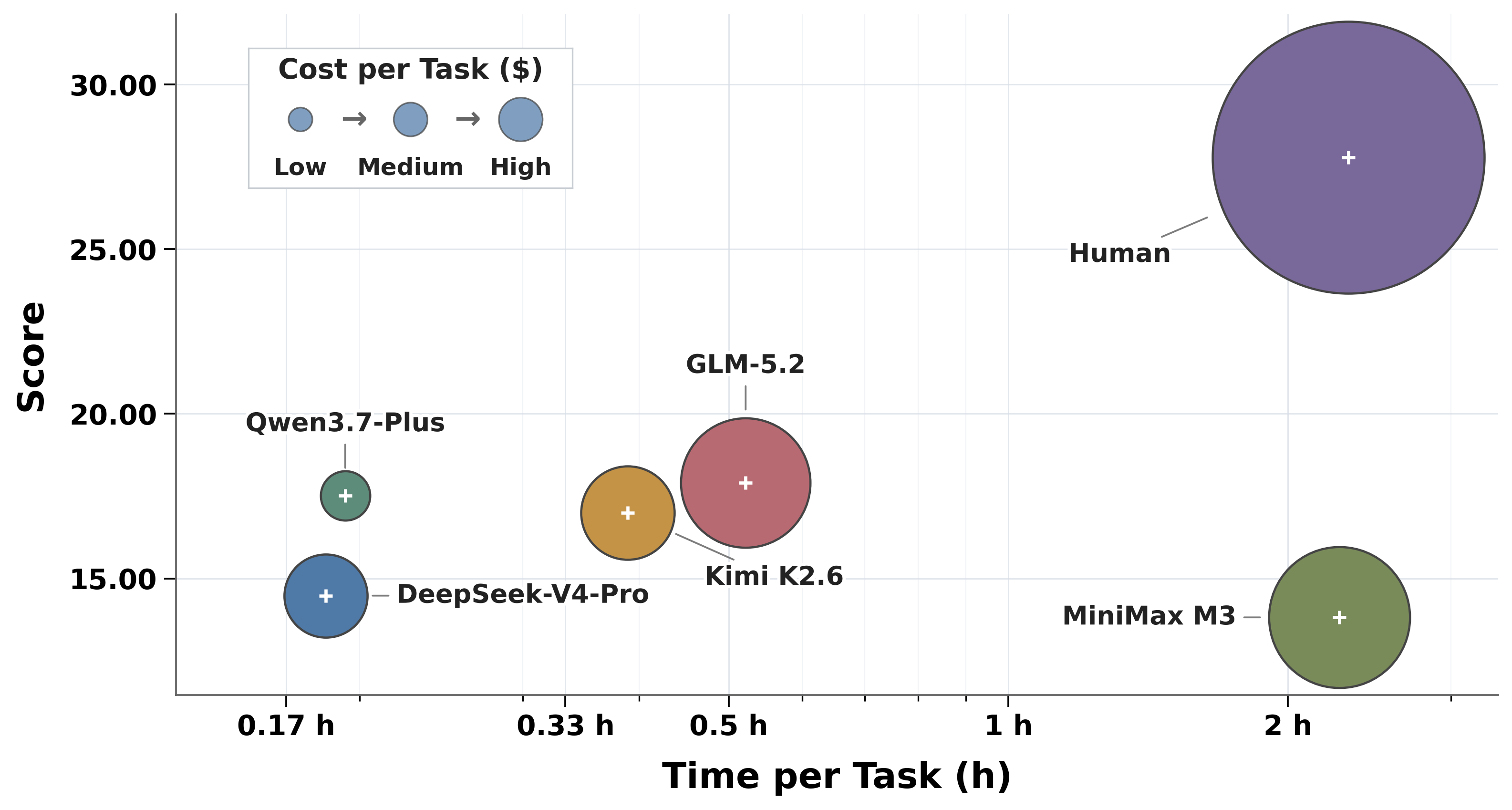}
    \caption{}
    \label{fig:homepage_comparison}
\end{subfigure}

\caption{
Overview of OmegaUse-OfficeVal.
(a) The benchmark construction pipeline.
(b) Task-level comparison of human annotators and LLM agents in terms of
score, runtime, and cost; bubble size represents the average
monetary cost per task.
}
\label{fig:homepage_overview}

\end{minipage}
\endgroup

\clearpage
\pagestyle{firstpage}

\tableofcontents

\clearpage
\pagestyle{normalpage}

\section{Introduction}

Agents have evolved rapidly from natural-language interaction toward delivering work products for human users. A key promise of such agents is their ability to assist workers with everyday office-suite productivity tasks, particularly those involving documents, spreadsheets, presentations, and PDFs. Although these tasks are common and may appear routine, they often require long-horizon execution. Powered by frontier large language models (LLMs), coding agents have achieved notable commercial success and widespread impact. Beyond coding, it is widely anticipated that the next step for LLM agents is the shift from ``vibe coding'' to ``vibe working''~\cite{chauhan2025vibeworking}: using LLM agents to complete long-horizon office-suite work. Such capabilities could substantially improve productivity and carry profound societal and economic implications. Realizing this potential, however, requires benchmarks that measure whether agents are actually capable of carrying out such office-suite workflows at an acceptable cost.

Existing benchmarks capture important aspects of this emerging landscape but leave several key gaps. General-purpose productivity benchmarks such as GDPVal~\cite{gdpval}, the Remote Labor Index (RLI)~\cite{rli}, and Agents' Last Exam (ALE)~\cite{ale} ground tasks in broad and complex professional workflows. However, they release only task subsets or restrict access to reference outputs, and they typically associate economic value with broad occupational categories rather than individual tasks. Moreover, these benchmarks are not specifically designed around office-suite workflows and often do not provide complete task-level economic annotations. Office-automation benchmarks such as OfficeBench~\cite{officebench}, OdysseyBench~\cite{odysseybench}, SpreadsheetBench~\cite{spreadsheetbench}, and PPT-Eval~\cite{ppteval} evaluate agents on document operations, spreadsheet manipulation, presentation editing, or standardized office proficiency. Nevertheless, existing office benchmarks generally focus on bounded operations, synthetic workflows, or standardized proficiency exercises. Most of them are not long-horizon tasks, and they do not provide economic annotations. GUI-agent benchmarks are likewise dominated by short-horizon tasks and are commonly evaluated according to whether an agent reaches a predefined environment state. Even recent long-horizon benchmarks, such as OSWorld 2.0, remain centered primarily on GUI interactions and process-level evaluation rather than the quality and usability of the final deliverable. As a result, current benchmarks offer limited support for the question that matters most in practice: given a long-horizon office-suite task, can an agent deliver an artifact that is correct and usable, and how much human labor or market value does that deliverable represent?

\begin{figure*}[t]
    \centering
    \includegraphics[width=0.98\linewidth]{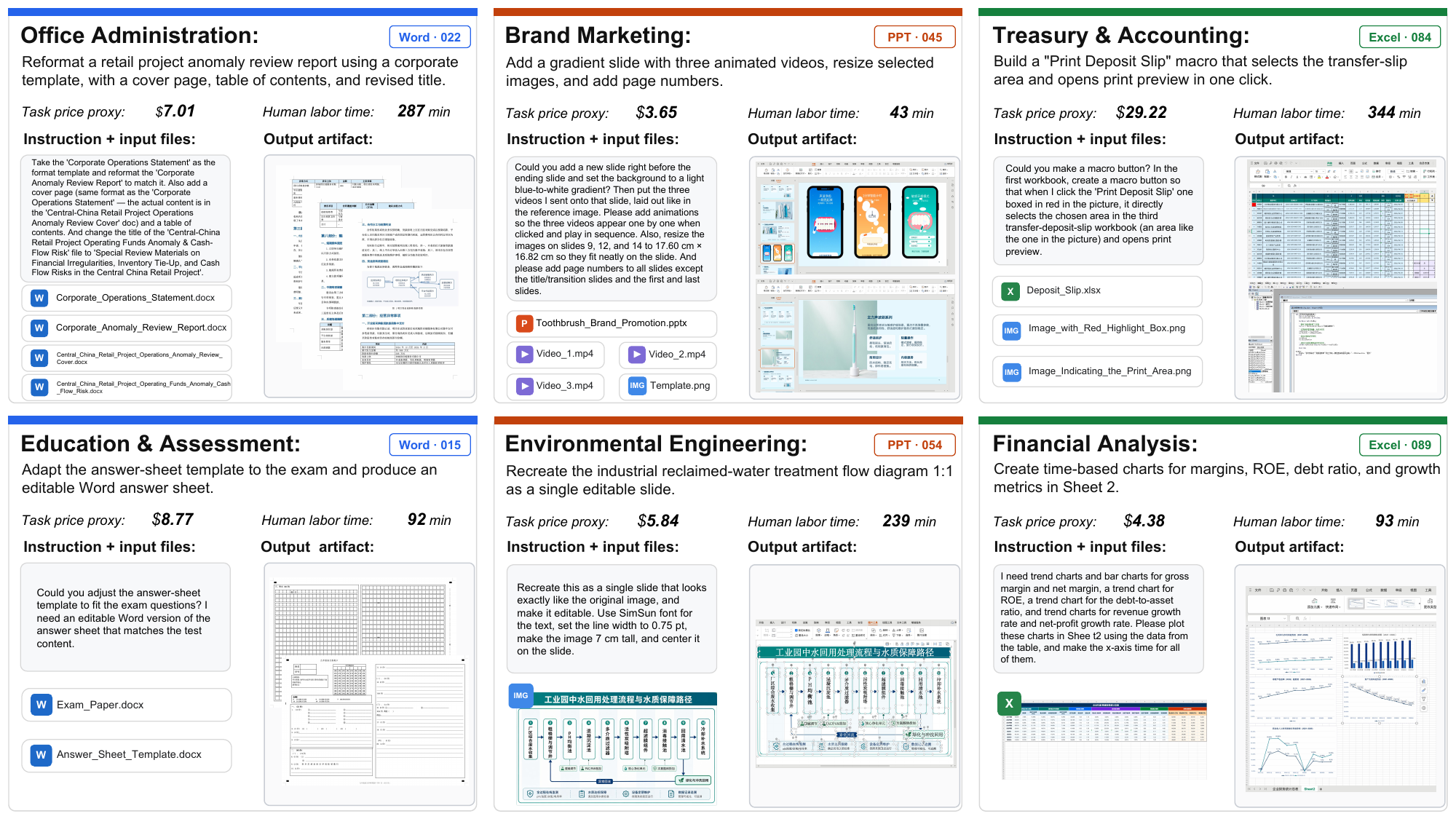}
    \caption{Task Example of OmegaUse-OfficeVal.}
    \label{fig:taskexample}
\end{figure*}

In this study, we introduce \textbf{OmegaUse-OfficeVal}, a benchmark for evaluating LLM agents on long-horizon office-suite tasks with task-level economic grounding. OmegaUse-OfficeVal contains 100 tasks derived from authentic office requests proposed by practitioners and adapted through a privacy-preserving process. Each task consists of a high-level user instruction and one or more input files, and requires the agent to produce deliverables, reflecting a practical setting in which a worker is assisted by an agent that takes over specific office-suite workflows. On average, the tasks in OmegaUse-OfficeVal require 2.32 hours of human labor to complete. Six illustrative examples selected from OmegaUse-OfficeVal tasks are shown in Figure~\ref{fig:taskexample}.

The construction of OmegaUse-OfficeVal is guided by three design principles. First, the benchmark should reflect authentic office productivity requirements originating from real workplace scenarios. The tasks should be meaningful while retaining clear objectives and concrete deliverables that could typically be completed by an entry-level office worker, assistant, or intern. Their instructions should also reflect how users naturally formulate requests in professional settings, rather than being expressed as artificial sequences of low-level operations. Second, unlike GDPVal and RLI, which aim to cover broad categories of economically valuable work, OmegaUse-OfficeVal focuses on the economic value of each individual task. This task-level perspective enables direct comparisons between human labor costs and LLM inference costs. Task-level economic grounding also enables value-weighted evaluation, complementing conventional analyses based on average task performance. Third, we include only long-horizon tasks. Although short and simple office tasks are also important, we deliberately select challenging, long-horizon tasks to probe the frontier capabilities and future potential of LLM agents.

As a result of these principles, a distinctive feature of OmegaUse-OfficeVal is that every task is paired with two economic signals. Human labor time records the time required by human annotators to complete the task without LLM assistance, collected under a human incentive protocol. Task price proxy estimates the market price of completing the task, using explicit price signals from practitioners when available and a consistency-based aggregation of expert estimates otherwise. To make evaluation reproducible and stable as LLMs evolve, we  avoid human and LLM-as-judge scoring and instead adopt code-based verifiers. Each task is verified through fine-grained rubrics that reward completed requirements and penalize unintended damage that increases the user's repair burden; these rubrics are translated into executable code and calibrated through an iterative human–code discrepancy resolution process.

We evaluate several LLMs on OmegaUse-OfficeVal, including GLM-5.2, Kimi K2.6, DeepSeek-V4-Pro, MiniMax M3, and Qwen3.7-Plus. We additionally include a human baseline for comparison. The results show that current LLM agents make meaningful but limited progress: the best LLM achieves a score of 17.91, compared with 27.79 for the human baseline. Value-weighted evaluation further reveals that the model with the highest average score is not necessarily the one that captures the most economically important tasks. Performance also varies with human labor time, demonstrating that long-horizon office-suite work remains a challenging and heterogeneous setting for LLMs. At the same time, all evaluated LLMs are cheaper and faster than human workers, although they have not yet approached human-level deliverable quality.
We summarize our main contributions as follows:
\begin{itemize}
    \item We introduce {OmegaUse-OfficeVal}, a benchmark comprising 100 long-horizon office-suite workflows. We release the complete benchmark assets with additional information available at \url{https://omegause-officeval.github.io}.
    \item We provide task-level economic grounding through human labor time and task price proxy annotations. We further develop a code-based verification protocol with usability checks and fine-grained rubrics to ensure stable evaluation.
    \item We evaluate a range of LLMs against human baselines, showing that evaluated LLMs are faster and cheaper but still lag behind humans in deliverable quality. 

\end{itemize}

\section{Related Work}\label{sec:related}
OmegaUse-OfficeVal is related to three lines of prior work: productivity-agent benchmarks, office-automation benchmarks, and computer-use benchmarks. We briefly survey related benchmarks here and defer a more comprehensive discussion to Appendix~\ref{app:more_related_work}. In general, OmegaUse-OfficeVal is distinguished by three points: a focus on \emph{long-horizon} office-suite workflows, \emph{task-level} economic grounding, and a fully \emph{open} benchmark release.

\subsection{Productivity-Agent Benchmarks}
A growing body of work evaluates LLM agents on realistic professional tasks with links to economic value. General-purpose efforts such as GDPVal~\cite{gdpval}, the Remote Labor Index (RLI)~\cite{rli}, and Agents' Last Exam (ALE)~\cite{ale} construct tasks from real or realistic professional workflows and assess whether agents can produce deliverables comparable in quality to those created by human professionals. Domain-specific benchmarks further target software engineering~\cite{swelancer}, enterprise SaaS workflows~\cite{workarena,workarenapp}, simulated companies~\cite{theagentcompany}, and high-value professional services~\cite{apex,apexagents}. Relative to this line of work, OmegaUse-OfficeVal offers two distinctive advantages. First, in terms of \emph{openness}, it releases the full benchmark assets, including task instructions, input files, rubrics, verifier code, and economic annotations, whereas benchmarks such as GDPVal, RLI, and ALE release only task subsets or gate access to reference outputs~\cite{gdpval,rli,ale}. Second, in terms of \emph{task-level economic grounding}, OmegaUse-OfficeVal attaches per-task human-labor-time and price-proxy annotations to every task, whereas these benchmarks provide economic grounding only at the aggregate level, for restricted public subsets, or implicitly through workflow realism~\cite{gdpval,rli,ale,swelancer}.

\subsection{Office-Automation Benchmarks}
A closely related work studies office-automation productivity. OfficeBench~\cite{officebench} evaluates multi-application workflows, and OdysseyBench~\cite{odysseybench} extends this direction toward longer office workflows. Other benchmarks target specific facets of office productivity, such as spreadsheets~\cite{spreadsheetbench}, presentations~\cite{pptc,ppteval,slidesbench}, and standardized office proficiency~\cite{mindthegap}. Two limitations recur across this line of work. First, most of these studies are characterized by application count, action steps or dialogue length rather than by \emph{measured human labor time}~\cite{docops,odysseybench}. In contrast, OmegaUse-OfficeVal targets long-horizon workflows that require roughly two hours of human labor on average. Second, these benchmarks generally do not provide per-task economic annotations. Thus, while they can assess whether an agent completes an office task, they offer limited support for analyzing performance in terms of the human effort or market value associated with each task.

\subsection{Computer-Use and GUI-Agent Benchmarks}
A parallel line of work evaluates GUI agents and computer-use agents (CUAs) in interactive, execution-based environments spanning the web~\cite{webarena,visualwebarena,mind2web,webvoyager} and operating-system or desktop settings~\cite{osworld,windowsarena,macosworld}. Recent work has also begun to develop general-purpose GUI agents for autonomous task execution across platforms \cite{omegause}. Correspondingly, benchmarks in this area have expanded to operating-system and desktop environments, enabling the evaluation of GUI agents on office-suite tasks. Most such benchmarks are short-horizon: their tasks are bounded interactions that can be completed in a small number of steps or a few minutes of human effort, which makes them well suited for evaluating perception, grounding, and short-term planning. OmegaUse-OfficeVal differs in two main respects. First, it does not prescribe or privilege a particular execution trajectory: agents may complete tasks through GUI interactions, scripts, or hybrid strategies, and evaluation targets the final office deliverable, including its task completion and usability, rather than the trajectory followed. Second, it focuses on long-horizon office-suite workflows rather than short-horizon interactions.

A notable recent exception is OSWorld 2.0~\cite{osworld2}, which reframes computer-use evaluation around long-horizon workflows and is therefore related to OmegaUse-OfficeVal. OSWorld 2.0 comprises 108 tasks whose median task takes a skilled human roughly 1.6 hours of active operation. Nevertheless, the two benchmarks differ fundamentally in their evaluation target. OSWorld 2.0 remains a computer-use benchmark centered on GUI operation, asking whether the agent reaches the expected environment state. OmegaUse-OfficeVal instead evaluates the final office deliverable itself. The two benchmarks also differ in economic grounding: OSWorld 2.0 reports human-annotated completion times and an aggregate economic-coverage analysis based on occupation-level mappings that characterize the benchmark as a whole, whereas OmegaUse-OfficeVal pairs each long-horizon task with task-level economic annotations, including human labor time and price proxies. This design enables value-aware analysis of agent performance.

\section{OmegaUse-OfficeVal Benchmark}

We first describe the structure and statistics of OmegaUse-OfficeVal. Each task provides a high-level user instruction and input files, and requires the agent to produce a final artifact. Evaluation focuses on the quality and correctness of the final artifact. The resulting dataset comprises 100 tasks.

\subsection{Dataset Description}

We describe the contents of OmegaUse-OfficeVal. Representative examples from the dataset are shown in Figure~\ref{fig:taskexample}. Each task consists of four main components:
\begin{itemize}
    \item \textbf{Instruction}: a textual description of the user request and the specific requirements to be satisfied.
    \item \textbf{Input files}: a collection of files required to complete the task specified by the instruction.
    \item \textbf{Human labor time}: the recorded time required by human workers to complete the task.
    \item \textbf{Task price proxy}: a task-level price signal based on an explicit price when available and otherwise obtained through expert estimation.
    \item \textbf{Code verifier}: executable evaluation code derived from the instruction and rubric, which assigns a score to an artifact produced by an LLM agent.
\end{itemize}

The input files are multimodal. In addition to textual information, they may contain images and videos, either as standalone files or embedded within word-processing documents, presentations, and spreadsheets. Note that the dataset records task price proxies in Chinese yuan (CNY). For better readability, we report them in U.S. dollars (USD) throughout the paper, using an exchange rate of 1 CNY = 0.14609 USD, computed as the 180-day average from January 19 to July 16, 2026.

\subsection{Task Statistics}

\begin{figure}[t]
\centering
\begin{minipage}[t]{0.4\linewidth}
\vspace{0pt}
\centering
\small
\captionof{table}{Distribution of input files and output artifacts.}
\label{tab:stat_modalities}
\begin{tabular}{lrr}
\toprule
\textbf{File type} & \textbf{Input files} & \textbf{Output artifacts} \\
\midrule
Video/Audio & 10 & 0 \\
Image       & 77 & 0 \\
DOCX        & 63 & 48 \\
PPTX        & 31 & 40 \\
XLSX        & 25 & 24 \\
PDF         & 14 & 3 \\
\midrule
\textbf{Total} & \textbf{220} & \textbf{115} \\
\bottomrule
\end{tabular}
\end{minipage}
\hfill
\begin{minipage}[t]{0.58\linewidth}
\vspace{0pt}
\centering
\includegraphics[width=\linewidth]{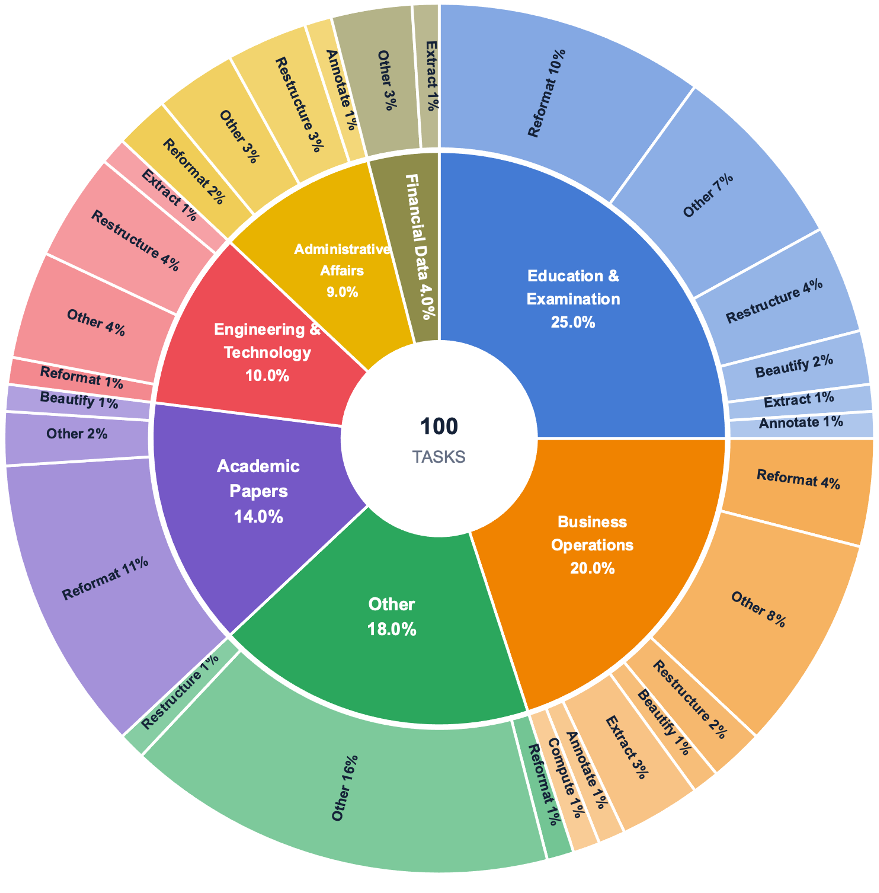}
\caption{Task distribution by domain and operation intent in OmegaUse-OfficeVal. The inner ring shows the domain distribution, and the outer ring shows the operation-intent distribution within each domain.}
\label{fig:task_distribution}
\end{minipage}
\end{figure}

Table~\ref{tab:stat_modalities} summarizes the distribution of modalities and file types across the input files and required output artifacts. OmegaUse-OfficeVal contains 100 tasks, and the distribution is not manually balanced. Instead, it reflects the observed distributional characteristics of office-suite tasks during the task collection process. The dataset includes images and videos as input files because real user requests are often concise, and many task-critical details are not explicitly stated in the textual instruction. Instead, they may be conveyed through images, videos, or documents. Therefore, OmegaUse-OfficeVal includes not only textual instructions, but also images, videos, and multiple types of Office files as task inputs.

\begin{figure}[t]
    \centering
\includegraphics[width=0.95\linewidth]{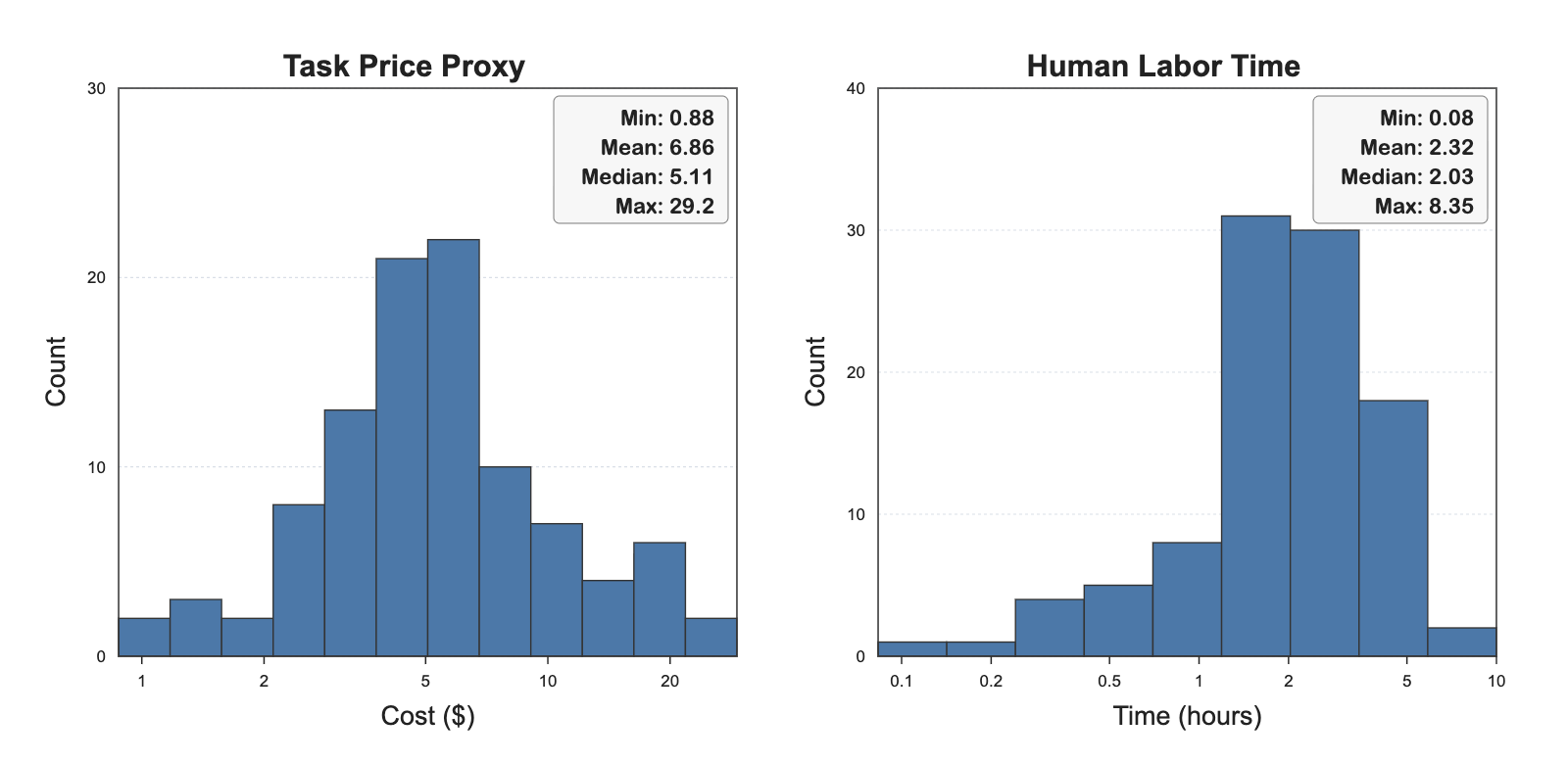}
    \caption{Distribution of task-level economically grounded annotations in OmegaUse-OfficeVal.}
    \label{fig:task-stat}
\end{figure}

Figure~\ref{fig:task-stat} summarizes the distribution of the two economically grounded annotations in OmegaUse-OfficeVal: task price proxy and human labor time. The task price proxy has a mean of \$6.86 and a median of \$5.11, indicating that the tasks correspond to moderately priced office work that would plausibly be assigned to a junior assistant or outsourced as a small office job. The human labor time distribution further highlights the long-horizon nature of the benchmark. The average task requires 2.32 hours of human labor, with a median of 2.03 hours and a maximum of 8.35 hours. This indicates that OmegaUse-OfficeVal contains tasks that require sustained execution. If such workflows could be completed reliably by LLM agents, they would represent a meaningful opportunity for improving office-suite productivity.

Figure~\ref{fig:task_distribution} shows the task distribution by domain and operation intent. The benchmark covers a broad range of office scenarios. The distribution is intentionally not uniform; instead, it reflects the types of office-suite requests observed during task collection. This distribution supports our goal of evaluating LLM agents on heterogeneous office-suite workflows. We provide the operation-intent and domain taxonomies in Appendix~\ref{sec:task_taxonomy}.

\begin{figure*}[t]
    \centering
    \includegraphics[width=0.98\linewidth]{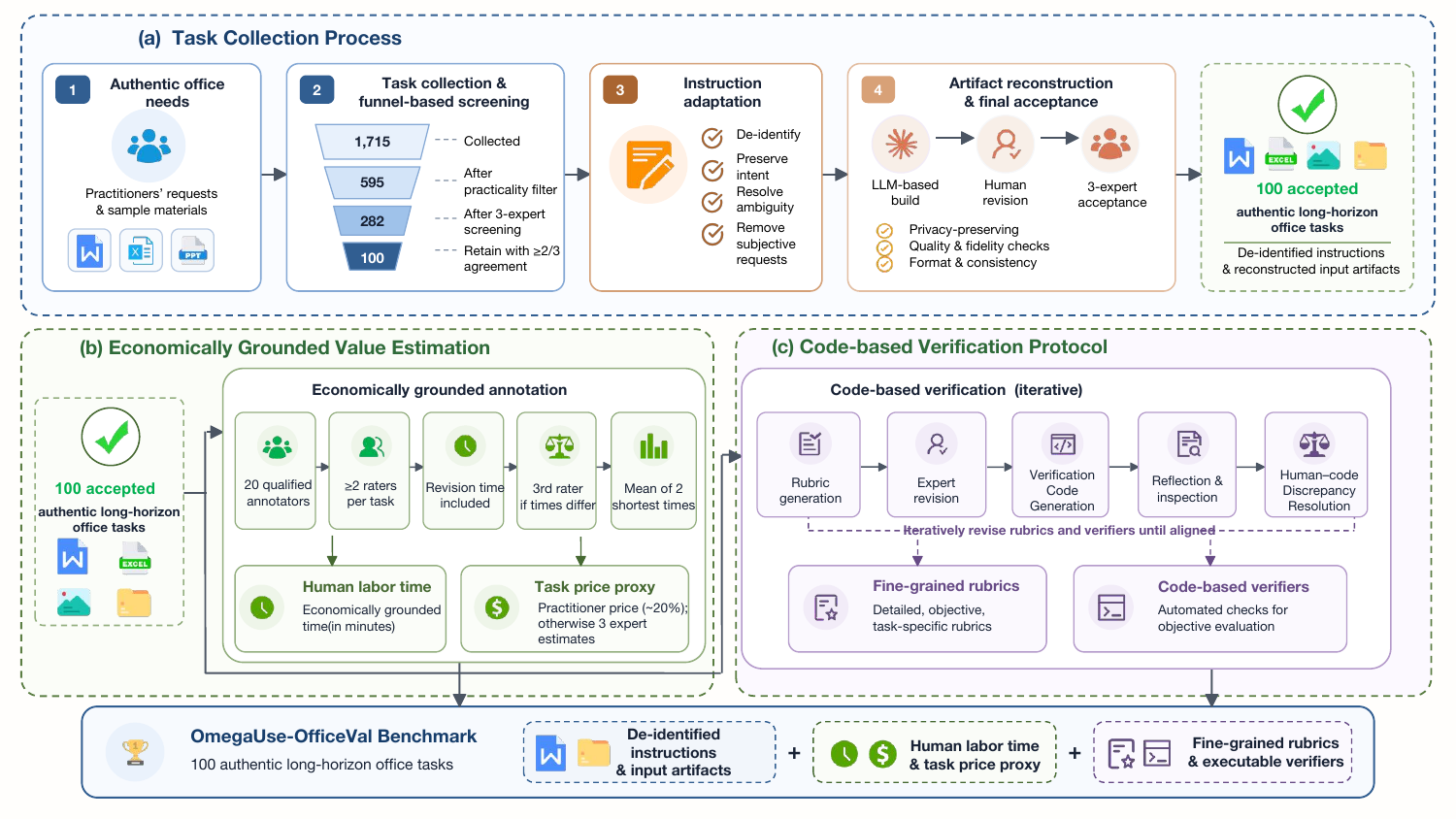}
    \caption{Task Construction Pipeline for OmegaUse-OfficeVal.}
    \label{fig:pipeline}
\end{figure*}

\section{Task Construction Pipeline}

As shown in Figure~\ref{fig:pipeline}, OmegaUse-OfficeVal is constructed through a multi-stage pipeline consisting of task collection, economically grounded value estimation, and code-based verification. We begin with authentic office needs and practitioner-provided sample materials, filter them through expert screening, and apply privacy-preserving adaptation. We then annotate each accepted task with human labor time and a task price proxy, and construct fine-grained rubrics and code-based verifiers for evaluation.

\subsection{Task Collection Process}

We first describe the task collection process for OmegaUse-OfficeVal. We recruit practitioners to propose realistic office-suite tasks grounded in their everyday workplace workflows and representative sample materials. The proposed tasks are intended to reflect work typically assigned to a junior worker, assistant, or intern, while still requiring concrete deliverables and nontrivial execution.

We adopt a funnel-based screening process to progressively filter candidate tasks. We initially collect 1,715 practitioner-proposed tasks. Experts first filter these tasks according to whether they are grounded in real office needs, have clearly specified task descriptions, and define clear expected deliverables. This stage yields 595 tasks. Next, three senior experts independently assess whether each task is sufficiently nontrivial and long-horizon, while remaining feasible for humans to complete under normal working conditions. A task is retained only when at least two of the three experts agree that it satisfies these criteria. This screening step yields a pool of 282 candidate tasks, from which we curate the final set of 100 tasks included in the benchmark.

After collecting the initial task descriptions, we preprocess the task instructions through privacy-preserving adaptation and expert review. Instructions are rewritten to remove sensitive or identifying information while preserving the original user intent, constraints, and task-critical details. Expert reviewers then inspect all instructions to resolve ambiguities and clarify missing information when such information is necessary for task completion. To ensure task objectivity, we remove subjective requirements that cannot be evaluated reliably. At the same time, we preserve the original colloquial style and natural phrasing of user requests whenever possible.

Input files are reconstructed with LLM assistance based on the task requirements and practitioners' representative sample materials. These files are then manually revised by annotators to improve realism, remove privacy or copyright risks, and ensure that the resulting artifacts remain faithful to the intended task. Annotators check for duplicated paragraphs, residual sensitive information, and inconsistencies introduced during generation. Images and videos are reproduced separately and reviewed before being inserted into the files to ensure that they are realistic and contextually appropriate. Annotators also inspect the layout of text and visual content to ensure that figures are placed appropriately within the document and do not overlap, float incorrectly, or obstruct surrounding content. They further examine document structure, formatting, section organization, and image-text layout to ensure that the files resemble authentic office materials and that no task-relevant information is lost during adaptation.

Finally, the revised and de-identified tasks undergo a final expert acceptance review. Three senior experts verify that each task poses no privacy risk, can be executed normally, and still conforms to the language, structure, and artifact format of real office scenarios. A task is included in OmegaUse-OfficeVal only when all three experts agree that it satisfies these acceptance criteria.

\subsection{Economically Grounded Value Estimation}
A distinctive feature of OmegaUse-OfficeVal is that each long-horizon task is annotated with task-level economic signals. We use two complementary indicators: \textbf{human labor time}, which measures the recorded time required by recruited annotators to complete the task without LLM assistance, and \textbf{task price proxy}, which estimates the market price of completing the task. Together, these indicators allow us to analyze LLM performance not only by task completion, but also in terms of the human effort and economic value associated with each task.

\subsubsection{Human Labor Time.}

We carefully collect human labor time for each task which measures the time required for a junior worker to complete a task without LLM assistance. We recruit 20 annotators to complete the tasks. Each annotator first undergoes an interview and completes sample tasks to verify that they have the basic skills required for office-suite work. We assign at least two annotators to each task; when their completion times differ substantially, we assign a third annotator to complete the same task. To encourage annotators to work efficiently while maintaining deliverable quality, we design a quality-gated incentive mechanism. For each task, we compute the human labor time as the average of the two shortest valid completion times. This aggregation reduces the influence of unusually slow attempts and provides a robust estimate of human completion time under quality-controlled conditions. The full incentive mechanism and human labor-time computation procedure are described in Appendix~\ref{apx:compute_human_labor_time} and Algorithm~\ref{alg:human_time_incentive}.

\subsubsection{Task Price Proxy.}

As a complement to human labor time, we annotate each task with a task price proxy, defined as an estimated market price for completing the task. During data collection, a small subset of tasks, about 20\%, were associated with high-confidence explicit price signals provided by practitioners, based on their prior experience of similar tasks previously outsourced on online platforms. Unlike human labor time, these value labels reflect users' perceived task importance or willingness to pay, and therefore capture a different aspect of the task's economic value in real-world productivity scenarios.

Because explicit price labels are sparse, we adopt a hybrid strategy that combines explicit price signals with expert estimates. This design grounds the economic signal in real user demand when available, while providing complete coverage across all tasks. Specifically, tasks with explicit price signals use those values directly as their task price proxy. For tasks without explicit price information, we introduce an expert estimation process: three experts with relevant domain experience independently estimate the task value based on the corresponding occupational category, task complexity, and typical labor compensation.

Thus, the task price proxy for each task is derived from one of two sources:
\begin{itemize}
    \item \textbf{Explicit price}: used directly when a task price is available.
    \item \textbf{Estimated price}: used when no explicit price is available.
\end{itemize}

For estimated prices, we aggregate the three expert annotations through a consistency-based procedure. Specifically, we first sort the three estimates as $\min \leq \mathrm{mid} \leq \max$, and compute the two adjacent gaps: $A=\max-\mathrm{mid}$ and $B=\mathrm{mid}-\min$. If $A/B \geq 2$, the maximum estimate is treated as a clear outlier and discarded, and we use the mean of the minimum and middle estimates. If $B/A \geq 2$, the minimum estimate is treated as a clear outlier and discarded, and we use the mean of the middle and maximum estimates. Otherwise, the three estimates are considered sufficiently consistent, and we use their mean as the final task price proxy. This strategy preserves the real-world grounding of explicit price signals while reducing the influence of extreme expert estimates.

\subsection{Code-based Verification Protocol}\label{sec:code-based-verification}

We adopt a deterministic code-based verification protocol to evaluate the completion quality of each task. OmegaUse-OfficeVal evaluates the final output files rather than enforcing a fixed execution trajector, where multiple valid workflows can lead to the same deliverable.
Two common alternatives to code-based verification are human evaluation and LLM-as-judge evaluation. Human evaluation, used as the primary evaluation mechanism in RLI and GDPVal, can assess complex and open-ended artifacts, but it incurs substantial cost, high evaluation latency, evaluator subjectivity, and limited reproducibility. These factors make it difficult to scale evaluation across many models and agent configurations.
Another alternative is to use a frontier model to judge the quality of generated artifacts, commonly referred to as LLM-as-judge evaluation.
For benchmark construction, we view two issues as particularly problematic. First, as frontier LLMs rapidly improve, the choice of judge model becomes unstable. Second, if the judge model is changed in future evaluations, previously reported scores may no longer be directly comparable.
To mitigate these limitations, OmegaUse-OfficeVal uses code-based verifiers as the primary evaluation mechanism. We build the verification protocol through an iterative process.

\subsubsection{Rubric Generation with Expert Revision.}
 For each task, we first construct fine-grained scoring rubrics with LLM assistance and expert revision. Each rubric contains two dimensions: a \textbf{usability rubric}, which checks whether the delivered artifact is valid, openable, editable, and practically usable, and a \textbf{task-completion rubric}, which rewards completed task-specific requirements and penalizes unintended changes that increase the user's repair burden. Note that we evaluate the artifact from the perspective of a user who receives the file, measuring both how much useful work the agent has completed and whether the delivered artifact remains practically usable with minimal additional human repair.

\subsubsection{Verification Code Generation and Human-Code Discrepancy Resolution.}  We then convert the expert-revised rubrics into executable verification code with the assistance of coding agents. The generated verifier is inspected by experts, revised through LLM reflection, and further calibrated through human-code discrepancy resolution, where expert scores and verifier scores are compared at the level of individual rubric items. This process yields reproducible artifact-level evaluation while remaining closely aligned with expert judgments based on the original user requirements. We iterate this process until the code-based scores are sufficiently aligned with expert judgment. Details of the rubric design, verifier generation, and discrepancy-resolution process are provided in Appendix~\ref{app:code_based_verification_details}.

\subsubsection{Rubric Statistics and Evaluation  Scoring}

After establishing the code-based verification protocol, OmegaUse-OfficeVal contains 219 usability check items and 2,009 task completion rubric check items (see Table \ref{tab:rubric_statistics}).

\begin{table}[h]
    \centering
    \small
    \caption{Statistics of rubric check items.}
    \label{tab:rubric_statistics}
    \begin{tabular}{lcc}
        \hline
        Rubric dimension & Total check items & Average per task \\
        \hline
        Dim-1: usability rubric & 219 & 2.19 \\
        Dim-2: task completion rubric & 2,009 & 20.09 \\
        \hline
    \end{tabular}
\end{table}

We compute each task score with a two-stage procedure: an artifact must first pass all usability check items, after which its normalized task completion score is computed from weighted positive and negative rubric items. For each task $t$, let $\mathcal{U}_t$ denote the set of usability check items and $\mathcal{C}_t$ denote the set of task completion scoring items. Each usability check item $u \in \mathcal{U}_t$ returns a binary value $I_u \in \{0,1\}$, indicating whether the submitted artifact satisfies the check item. The task passes the usability rubric only if all usability check items are satisfied:
\[
U_t = \prod_{u \in \mathcal{U}_t} I_u .
\]

For task completion, each scoring item $c \in \mathcal{C}_t$ has a discrete weight $w_c$, where $w_c > 0$ for required task components and $w_c < 0$ for unintended errors or avoidable damage. Let $I_c \in \{0,1\}$ indicate whether scoring item $c$ is triggered. The raw task completion score is computed as
\[
S_t^{\mathrm{raw}} = \sum_{c \in \mathcal{C}_t} w_c I_c .
\]

Because positive requirements can be enumerated from the task instruction whereas negative failure modes cannot be exhaustively enumerated, we clip the raw score at zero to avoid assigning negative task completion:
\[
S_t^{\mathrm{clip}} = \max(0, S_t^{\mathrm{raw}}).
\]

The maximum attainable positive score is
\[
S_t^{+} = \sum_{c \in \mathcal{C}_t: w_c > 0} w_c .
\]

The final normalized score for task $t$ is then defined as
\[
\mathrm{Score}(t) =
U_t \cdot \frac{S_t^{\mathrm{clip}}}{S_t^{+}}.
\]

Thus, if the artifact fails any usability check item, $U_t = 0$ and the task receives a score of zero. Otherwise, the score is determined by the weighted task completion rubric, with negative items penalizing unintended damage while the final normalized score is lower-bounded by zero.

\section{Experiments}

\subsection{Setup}

\subsubsection{Models and Human Baseline.}
We evaluate LLMs from several model families, including GLM-5.2~\cite{glm5report,glm52blog}, Kimi K2.6~\cite{kimik25,kimik26blog}, DeepSeek-V4-Pro~\cite{deepseekv4pro}, MiniMax M3~\cite{minimaxmsa,minimaxm3blog}, and Qwen3.7-Plus~\cite{qwen3report,qwen37plusblog}. We also include the human annotators from the human labor time annotation process as a reference baseline. Since each task is completed by at least two human annotators, we use the highest-scoring human submission for each task as the human delivered artifact. We use this best human submission because human deliverable quality can vary substantially across annotators, whereas LLM evaluations are run under a fixed scaffold and are therefore more reproducible for a given model and task. Human annotator baseline represents a best junior-worker reference level and provides a point of comparison for assessing the current capability gap between LLM agents and human office workers. More detailed information about the setup can be found in Appendix \ref{apx:exp_setting}.

\subsubsection{Evaluation Metrics}\label{sec:evaluation_metrics}

We report three metrics in the experimental evaluation: score, time-weighted score, and price-weighted score. The overall score measures aggregate score across all tasks using the task score defined in Section~\ref{sec:code-based-verification}. The time-weighted score weights each task by its recorded human labor time, while the price-weighted score weights each task by its task price proxy. These two value-weighted metrics allow us to evaluate not only how well a model performs on average, but also how much human labor time or market value it captures across the benchmark. The detailed definitions of these metrics are provided in Appendix~\ref{apx:evaluation_metrics}.

\begin{table*}[t]
\centering
\small
\caption{Overall performance and efficiency on OmegaUse-OfficeVal. The best non-human result is underlined.}
\label{tab:model_performance_efficiency}
\begin{tabular}{lccccc}
\toprule
\textbf{Model} &
\textbf{Score} &
\textbf{Time-Weighted Score} &
\textbf{Price-Weighted Score} &
\textbf{Cost/Task (USD)} &
\textbf{Time/Task (hours)} \\
\midrule
Human       & 27.79 & 54.45 & 144.61 & \$6.8560 & 2.324 \\
\hline
GLM-5.2      & \underline{17.91} & 30.13 & 93.51 & \$1.4823 & 0.521 \\
Qwen3.7-Plus & 17.51 & \underline{34.73} & \underline{106.50} & \$\underline{0.2152} & 0.193 \\
Kimi K2.6    & 17.00 & 31.29 & 95.96 & \$0.7719 & 0.389 \\
DeepSeek-V4-Pro & 14.48 & 27.14 & 79.33 & \$0.6111 & \underline{0.184} \\
Minimax M3   & 13.82 & 30.81 & 76.97 & \$1.7572 & 2.275 \\
\bottomrule
\end{tabular}
\end{table*}

\subsection{Main Results}
\label{sec:main_results}

Table~\ref{tab:model_performance_efficiency} summarizes the overall performance and efficiency of human annotators and LLMs on OmegaUse-OfficeVal. The human achieves a score of 27.79, substantially outperforming all evaluated LLMs, but remains far from perfect. This reflects the difficulty of the benchmark: the tasks are long-horizon and complex, and the scoring protocol penalizes not only missing requirements but also unintended changes or avoidable damage to the final deliverable.

Among the LLM agents, GLM-5.2 achieves the highest average score, with a score of 17.91, followed closely by Qwen3.7-Plus and Kimi K2.6. However, the ranking changes when task-level economic value is considered. Qwen3.7-Plus obtains the highest time-weighted and price-weighted scores among the evaluated models, suggesting that it performs relatively better on tasks associated with greater human labor time or higher task price. This demonstrates the importance of value-weighted evaluation: the model with the highest average completion quality is not necessarily the model that captures the most economically important tasks.

The efficiency results reveal different trade-offs across models. DeepSeek-V4-Pro has the lowest average runtime per task, while Qwen3.7-Plus has the lowest average cost per task and remains competitive in completion quality. GLM-5.2 achieves the best average score, but requires higher cost and runtime than several alternatives. Overall, the evaluated LLMs are substantially cheaper and faster than the human baseline under our scaffold, but still lag behind junior human workers in deliverable quality. These results indicate that long-horizon office-suite workflows remain challenging for LLMs, while also suggesting substantial room for future progress.

\begin{figure}[t]
    \centering
    \includegraphics[width=\columnwidth]{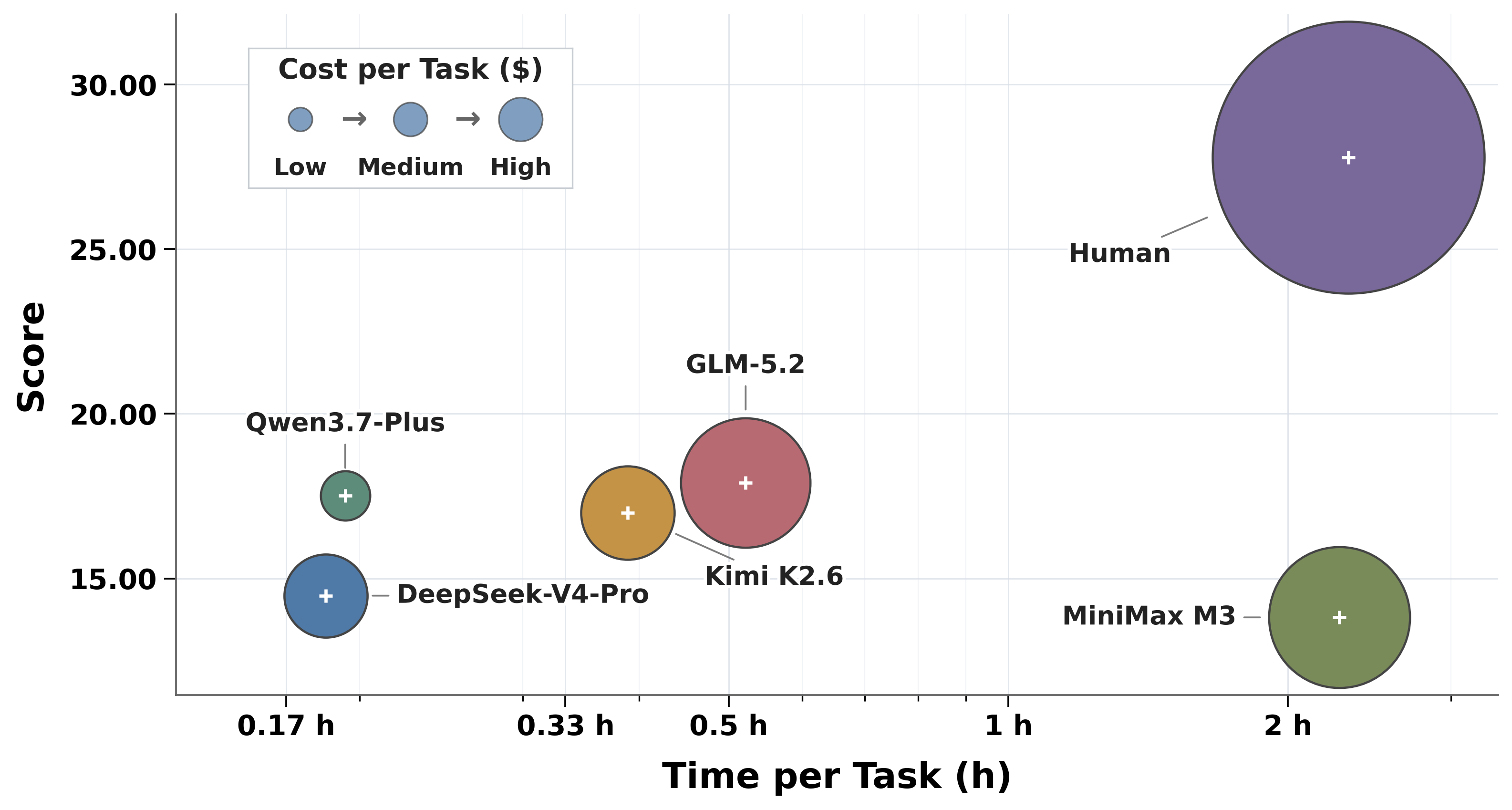}
    \caption{Score, runtime, and monetary cost across human annotators and LLM agents. The x-axis shows average time per task, the y-axis shows the score, and bubble size indicates average monetary cost per task. For human annotators, monetary cost is measured by the task price proxy; for LLM agents, it is measured by token cost.}
    \label{fig:time_score_cost_comparison}
\end{figure}

Figure~\ref{fig:time_score_cost_comparison} provides an illustrative view among overall score, runtime, and monetary cost. The desirable region is the upper-left corner with a smaller bubble, corresponding to high task quality, low time cost, and low monetary cost. Human annotators achieve the highest score, but require substantially more time and monetary cost per task. In contrast, all LLM agents are much cheaper and generally faster, but remain clustered at considerably lower scores. This indicates that current LLMs already provide efficiency advantages, but have not yet converted these advantages into human-level deliverable quality.

Figure~\ref{fig:time_score_cost_comparison} also reveals differences among LLMs. GLM-5.2 achieves the highest score among the models, but with higher runtime and cost than several alternatives. Qwen3.7-Plus is both low-cost and fast, but its completion quality still leaves substantial room for improvement. DeepSeek-V4-Pro is the fastest model, but lower runtime does not necessarily lead to better task completion. Overall, this figure suggests that the key challenge for future office-suite agents is not only to reduce cost and latency, but also to improve deliverable quality while preserving these efficiency advantages.

\subsection{Task Score Distribution}
\label{sec:score_dist}
In this section, we analyze the distribution of task scores across models and task groups. Additional breakdown analyses are provided in Appendix~\ref{app:additional_results}.

\begin{figure}[t]
    \centering
    \includegraphics[width=0.95\columnwidth]{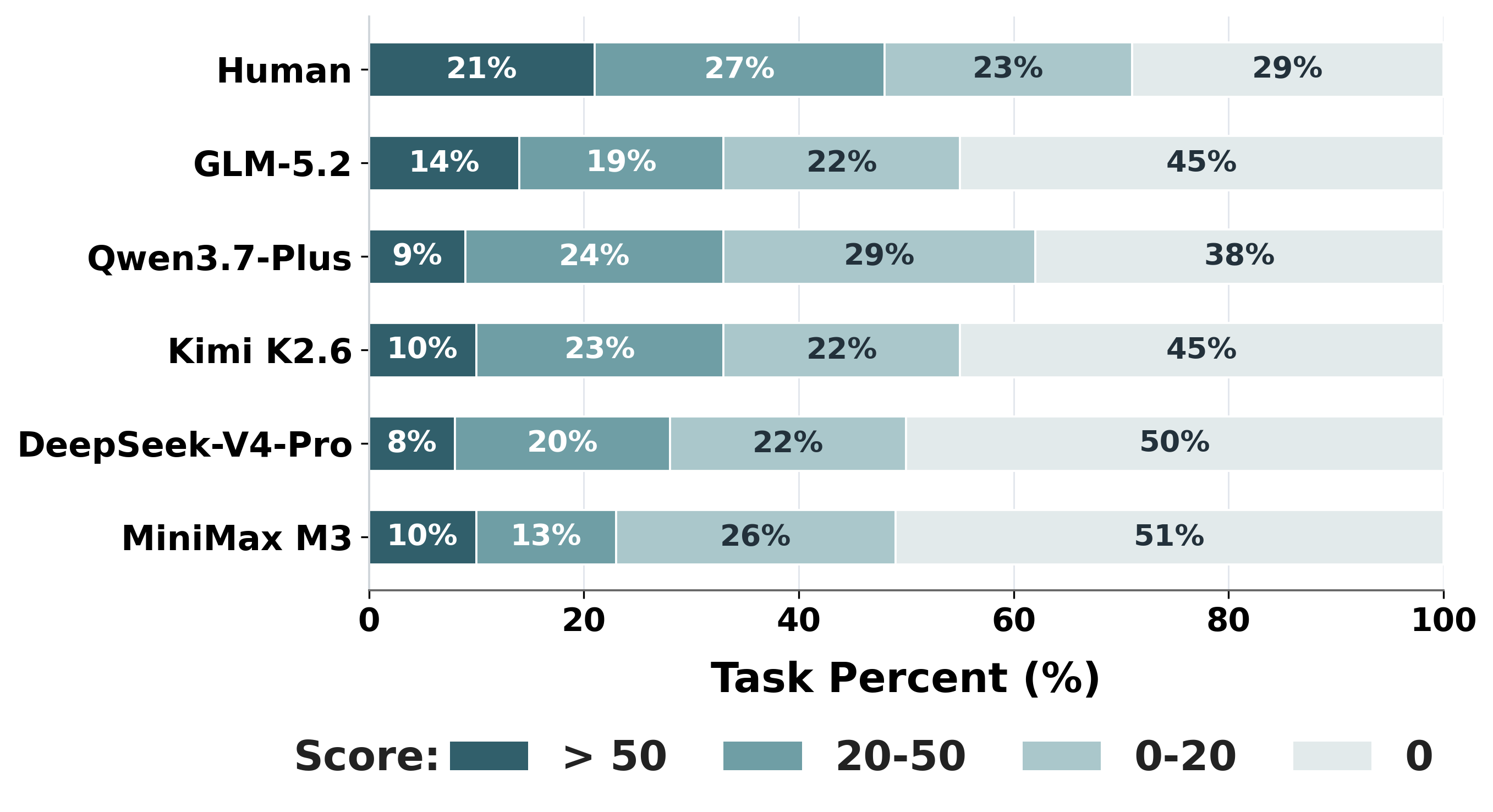}
    \caption{Distribution of task scores across human and LLMs. Tasks are grouped into four score buckets: zero, low score $(0,20]$, moderate score $(20,50]$, and high score $>50$.}
    \label{fig:task_score_distribution}
\end{figure}

Figure~\ref{fig:task_score_distribution} shows model performance by task-level score buckets. Human annotators have the largest share of high-score tasks, with 21\% of tasks scoring above 50, and the smallest share of zero-score tasks, at 29\%. This distribution confirms that the human baseline is not only better on average, but also more reliable across tasks: humans produce fewer complete failures and more substantially completed deliverables.

As we can see from Figure~\ref{fig:task_score_distribution}, among the LLM agents, GLM-5.2 has the largest fraction of high-score tasks, with 14\% of tasks scoring above 50, consistent with its highest average score in Table~\ref{tab:model_performance_efficiency}. Qwen3.7-Plus has the lowest zero-score rate among the LLM agents, at 38\%, and the largest share of low-to-moderate-score tasks when combining the $(0,20]$ and $(20,50]$ buckets. This suggests that Qwen3.7-Plus is relatively more robust at making partial progress. In contrast, DeepSeek-V4-Pro and MiniMax M3 have zero-score rates of 50\% and 51\%, respectively, indicating that a substantial fraction of office-suite tasks still result in complete failure under the current scaffold.

\begin{figure}[t]
    \centering
    \includegraphics[width=0.95\columnwidth]{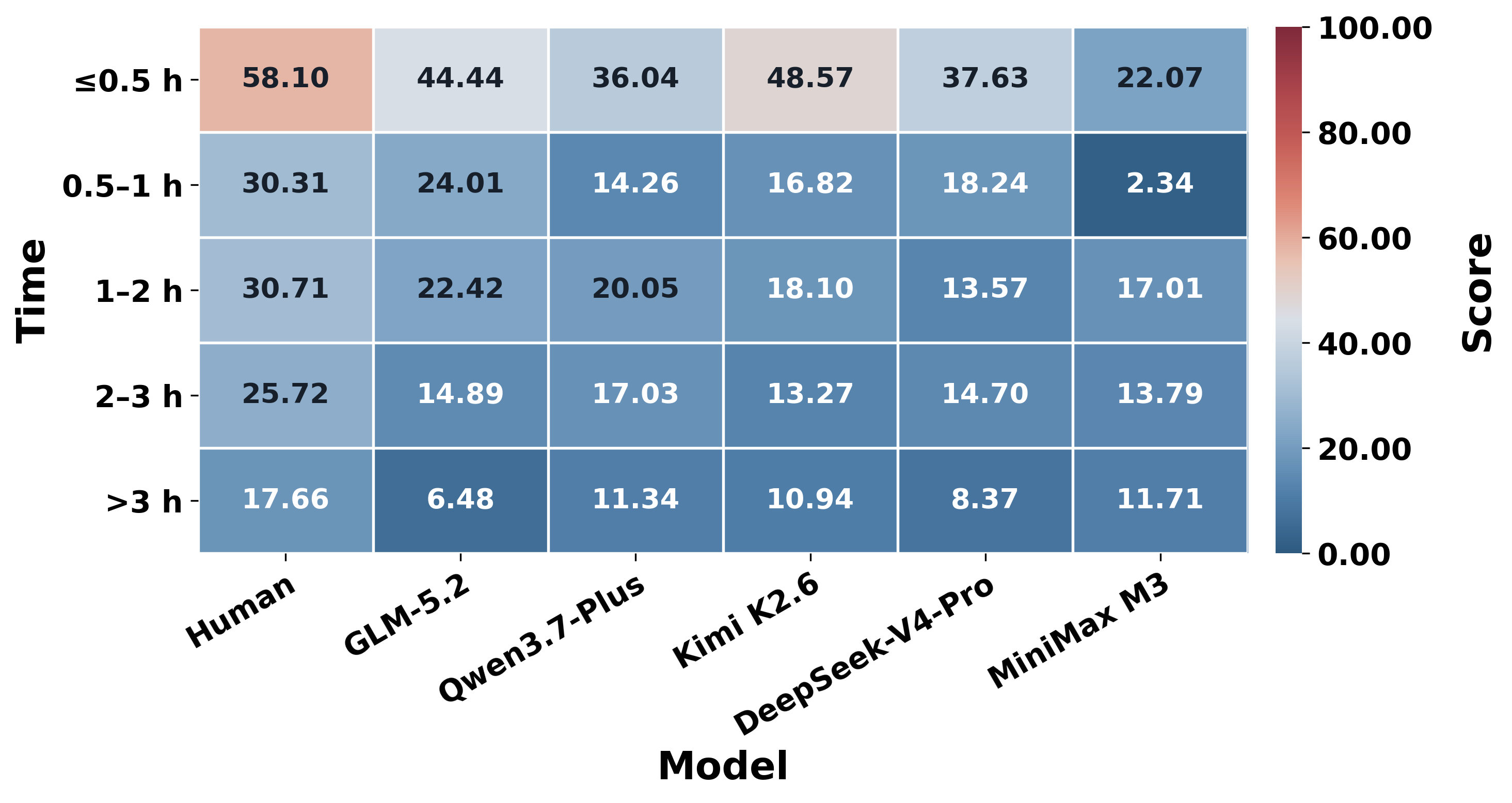}
    \caption{Task scores by human labor-time bucket. Tasks are grouped according to human labor time, and each cell reports the average score within the corresponding bucket.}
    \label{fig:time_score_heatmap}
\end{figure}

Figure~\ref{fig:time_score_heatmap} analyzes performance across task groups defined by human labor time. The results show a negative relationship between human labor time and score: tasks that require more human effort tend to receive lower scores from both human and LLMs. 
This pattern in Figure~\ref{fig:time_score_heatmap} suggests that human labor time is a useful indicator for task difficulty in OmegaUse-OfficeVal. Longer tasks typically require more sustained planning, more detailed file manipulation, and more opportunities for structural errors to accumulate. The decline is particularly pronounced for LLMs, indicating that current agents struggle to maintain task state and deliverable quality over extended office-suite workflows.

\subsection{Coding Actions vs. Computer-Use Actions on Office-Suite Tasks}

\begin{figure*}[t]
    \centering

    \begin{subfigure}[t]{0.48\textwidth}
        \centering
        \includegraphics[width=\linewidth]{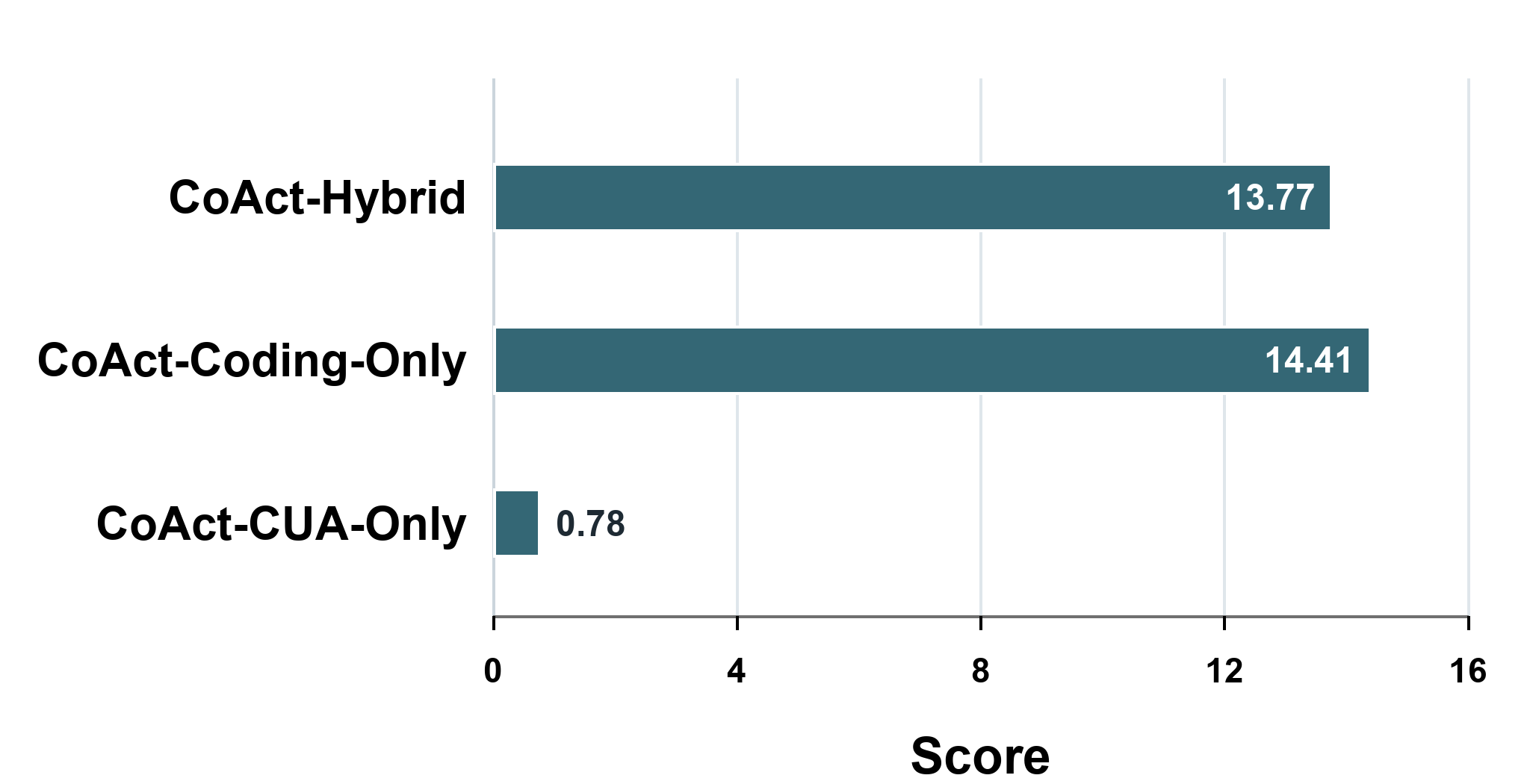}
        \caption{CoAct variants on OmegaUse-OfficeVal.}
        \label{fig:coact-officeval}
    \end{subfigure}
    \hfill
    \begin{subfigure}[t]{0.48\textwidth}
        \centering
        \includegraphics[width=\linewidth]{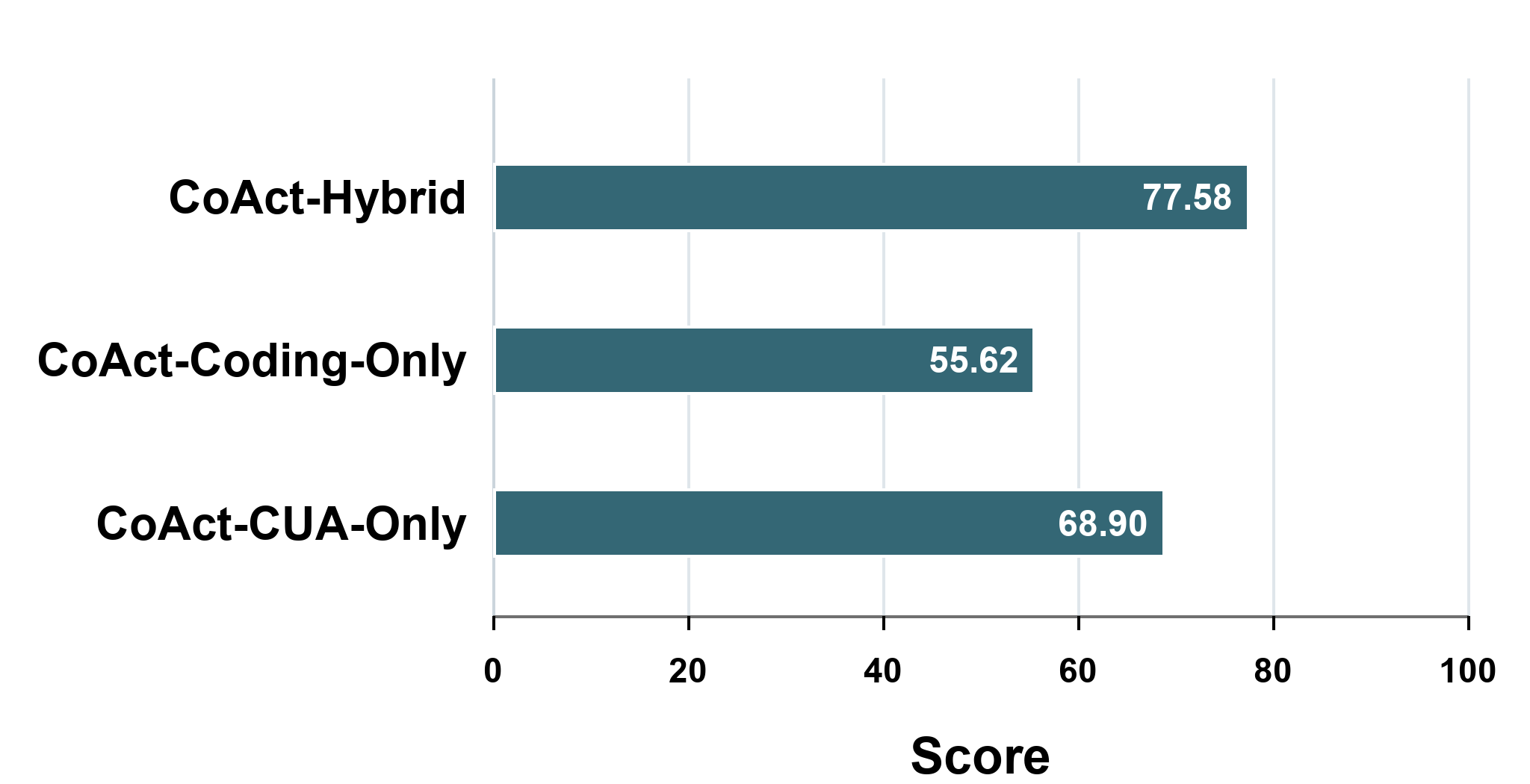}
        \caption{CoAct variants on OSWorld office-related tasks.}
        \label{fig:coact-osworld-office}
    \end{subfigure}

    \caption{
    Comparison of CoAct variants with K2.6 on OmegaUse-OfficeVal and OSWorld office-related tasks.
    }
    \label{fig:coact-officeval-osworld}
\end{figure*}

Recent work raises a question for long-horizon agents: should
office-suite tasks be solved through GUI interaction, through programmatic file
manipulation, or through a hybrid of the two? GUI-based computer-use agents offer
flexible interaction with visual application interfaces, but they can be brittle
and inefficient on long-horizon tasks with dense GUI elements, where small
grounding errors may accumulate over many steps \cite{song2026coact1}. By
contrast, coding agents can directly inspect and manipulate files and perform
precise data processing, but they may lack the interactive flexibility required
for visual inspection, layout refinement, or application-level operations.
CoAct-1 \cite{song2026coact1} addresses this tension by combining a
computer-use agent with a coding agent in a multi-agent system that dynamically
routes subtasks between GUI actions and coding actions.

To examine this trade-off in office-suite settings, we evaluate three CoAct
variants using K2.6 as the backbone model: {CoAct-Hybrid}, which combines
GUI and coding actions as proposed in CoAct-1 \cite{song2026coact1};
{CoAct-Coding-Only}, which disables GUI operation and permits only coding
actions; and {CoAct-CUA-Only}, which disables coding actions and relies
solely on computer-use actions. Figure~\ref{fig:coact-officeval-osworld}
compares their performance on OmegaUse-OfficeVal and on OSWorld office-related
tasks, including \texttt{libreoffice\_calc}, \texttt{libreoffice\_impress}, and
\texttt{libreoffice\_writer}. Because the two benchmarks differ in task
construction and evaluation protocol, we focus on the relative ranking of the
variants within each benchmark.

As shown in Figure~\ref{fig:coact-officeval-osworld}, the two benchmarks exhibit
different trends. On OSWorld office-related tasks, {CoAct-Hybrid} achieves
the highest score (77.58), outperforming both {CoAct-CUA-Only} (68.90)
and {CoAct-Coding-Only} (55.62). This result is consistent with the
motivation of CoAct-1: many OSWorld office tasks benefit from combining visual
application interaction with programmatic execution. The GUI agent can handle
frontend operations and visually grounded interactions, while the coding agent
can perform precise backend operations when appropriate.

OmegaUse-OfficeVal presents a different pattern. On this benchmark,
{CoAct-Coding-Only} obtains the highest score (14.41), slightly
outperforming {CoAct-Hybrid} (13.77), while {CoAct-CUA-Only}
performs poorly (0.78). This suggests that, for long-horizon office-suite tasks,
pure GUI interaction is often insufficient. GUI-only execution is vulnerable to
grounding ambiguity, long action chains, and cumulative errors. Coding actions,
in contrast, can be more reliable.  

These results should not be interpreted as evidence that coding agents can
replace computer-use agents in general. Rather, they suggest that, on
long-horizon office-suite tasks with strict artifact-level evaluation, coding
actions may provide important advantages over pure GUI interaction. At the same
time, the strong performance of {CoAct-Hybrid} on OSWorld office-related
tasks indicates that visual interaction remains valuable in settings that require
application-level operation or visual grounding. Understanding when to invoke
computer-use actions, when to rely on coding actions, and how to combine the two
effectively remains an important open problem for building reliable
office-suite agents.

\section{Conclusion}
In this study, we introduced OmegaUse-OfficeVal and conducted an experimental evaluation of LLM agents on the benchmark. OmegaUse-OfficeVal comprises 100 tasks, each paired with two economic signals, namely human labor time and a task price proxy, and verified through code-based verifiers built from fine-grained rubrics. By focusing on final office artifacts, the benchmark supports reproducible evaluation of whether agents can produce deliverables that are correct, usable, and practically valuable. Our experiments show that current LLM agents make meaningful progress on office-suite workflows, yet still remain substantially behind the human baseline in deliverable quality. We hope that OmegaUse-OfficeVal, together with its released assets, will serve as a rigorous testbed for tracking progress toward reliable ``vibe working.''

\bibliographystyle{ACM-Reference-Format}
\bibliography{refs}

\appendix

\section{More Related Work}
\label{app:more_related_work}
This appendix provides a comprehensive comparison between OmegaUse-OfficeVal and prior work along the three lines introduced in Section~\ref{sec:related}: productivity-agent benchmarks, office-automation benchmarks, and computer-use benchmarks. We describe the relevant benchmarks and analyze how OmegaUse-OfficeVal differs in task horizon, economic grounding, and openness.

\subsection{Productivity-Agent Benchmarks}
A growing body of work evaluates LLM agents on realistic productivity tasks with links to economic value. These efforts can be broadly divided into general-purpose productivity benchmarks and domain-specific productivity benchmarks. Relative to both categories, OmegaUse-OfficeVal offers two key advantages: greater openness and task-level economic grounding.

GDPVal~\cite{gdpval} and the Remote Labor Index (RLI)~\cite{rli} are representative general-purpose examples: both construct tasks from real or realistic professional workflows and assess whether agents can produce deliverables comparable in quality to those created by human professionals. Agents' Last Exam (ALE)~\cite{ale} similarly evaluates AI agents on long-horizon, economically valuable professional workflows in realistic desktop sandboxes, using hidden reference outputs and verifiable grading to measure whether agents can complete real-world work across many domains.

Compared with GDPVal, RLI, and ALE, OmegaUse-OfficeVal focuses on office-suite tasks rather than general-purpose jobs and offers two distinct advantages. First, in terms of openness, OmegaUse-OfficeVal provides a more complete public release than these broader productivity benchmarks. GDPVal and ALE release only subsets of their task pools, and ALE further gates access to reference outputs, while RLI releases only 10 public tasks and keeps its main evaluation set private. In contrast, OmegaUse-OfficeVal releases the full benchmark assets, including task instructions, input files, rubrics, verifier code, economic annotations, and additional task metadata, enabling fully reproducible evaluation and detailed task-level analysis. Second, in terms of task-level economic grounding, OmegaUse-OfficeVal provides per-task economic value annotations, whereas GDPVal, RLI, and ALE do not. GDPVal estimates task-level economic value from expert-estimated completion time and occupation-level wage statistics, but its public release does not expose task-level time or value annotations. RLI records project-level cost and human completion time from professional freelancers and releases such metadata only for its 10 public projects, while its 230-project official evaluation set remains private. Based on the public documentation, ALE frames its tasks as economically valuable professional workflows but, unlike OmegaUse-OfficeVal, does not provide explicit task-level economic annotations such as human labor time or task-price proxies.

More recent benchmarks extend this direction toward professional and enterprise workflows and are better characterized as domain-specific productivity or workplace-agent benchmarks rather than office-suite benchmarks. SWE-Lancer~\cite{swelancer} targets real Upwork software-engineering tasks with real payouts, covering both coding tasks and engineering-management decisions. TheAgentCompany~\cite{theagentcompany} evaluates agents in a simulated software-company environment, where they act as digital workers across code, files, web applications, chat, HR, finance, and project-management systems. WorkArena and WorkArena++~\cite{workarena,workarenapp} target enterprise SaaS workflows on ServiceNow, including form filling, ticket handling, knowledge-base search, service-catalog operations, and compositional workplace tasks. WorkBench~\cite{workbench} evaluates common business activities, such as email, calendar scheduling, CRM updates, project management, and website analytics, in simulated workplace databases and tools. Finally, APEX~\cite{apex} focuses on high-value single-turn professional tasks in investment banking, consulting, law, and medicine, while APEX-Agents~\cite{apexagents} extends this setting to long-horizon, cross-application professional-service tasks in investment banking, consulting, and corporate law.

These benchmarks also differ substantially in openness. APEX releases only a subset of development cases and relies on hidden held-out evaluation sets; SWE-Lancer releases a public evaluation split rather than the full underlying Upwork task assets; and WorkArena/WorkArena++ are open-source but depend on external enterprise software environments such as ServiceNow. WorkBench and TheAgentCompany provide more complete open-source simulated environments, while APEX-Agents reports an open release of tasks, rubrics, gold outputs, files, and infrastructure, although some tasks may still depend on external APIs or environment-specific resources.

Finally, these benchmarks vary in the extent to which they disclose task-level economic grounding. SWE-Lancer provides the strongest monetary grounding, as each software-engineering task is tied to a real Upwork payout, although the released public split does not expose the full underlying marketplace task assets. APEX and APEX-Agents are framed around economically valuable professional work and report expert-estimated task durations or professional-service settings, but they do not provide the fully released per-task price labels that OmegaUse-OfficeVal offers. WorkArena/WorkArena++ and WorkBench focus on realistic workplace or enterprise-software operations, but their tasks are grounded primarily by workflow realism rather than by explicit task-level monetary value or human-labor-time annotations. TheAgentCompany likewise evaluates consequential workplace tasks in a simulated company environment, but its economic grounding is implicit in the workplace setting rather than expressed through per-task human completion time or price fields. In contrast, OmegaUse-OfficeVal explicitly releases task-level economic annotations, including human labor time and task-price proxies, for all benchmark tasks.

\subsection{Office-Automation and Office-Suite Benchmarks}
A closely related line of work studies office automation and file-centric productivity tasks. These benchmarks are directly relevant to OmegaUse-OfficeVal because they also target practical office scenarios. Compared with this prior work, OmegaUse-OfficeVal is distinguished by two design choices: it focuses on long-horizon office workflows, and it provides task-level economic grounding.

OfficeBench~\cite{officebench} is among the first office-automation benchmarks, evaluating agents across multiple applications, including Word, Excel, PDF, email, and calendar, with an emphasis on multi-application workflows in a simulated office environment. OdysseyBench~\cite{odysseybench} extends this direction toward longer office workflows, requiring agents to extract essential information from extended interaction histories and coordinate multi-step reasoning across the same family of applications. Other benchmarks examine specific facets of office productivity: SpreadsheetBench~\cite{spreadsheetbench} targets real-world spreadsheet manipulation drawn from online forums and adopts an online-judge-style evaluation, while presentation-focused benchmarks such as PPTC~\cite{pptc} and PPT-Eval~\cite{ppteval} study PowerPoint task completion and editing. Closest in spirit to our verification philosophy, Mind the Gap/OfficeEval~\cite{mindthegap} constructs an office-proficiency evaluation from a standardized certification exam, comprising a large number of machine-gradable criteria across Word, Excel, and PowerPoint. Additional single-modality or workflow-specific efforts on document editing, slide generation, and application operation are also related but narrower in scope, including DocOps~\cite{docops} and SheetCopilot~\cite{sheetcopilot}.

With respect to task horizon, existing office and document benchmarks characterize difficulty mainly through application count, action steps, dialogue length, rubric density, or agent runtime, rather than through measured human labor time. OfficeBench~\cite{officebench} emphasizes multi-application planning, but its tasks are framed around bounded operation chains within a simulated office environment. OdysseyBench~\cite{odysseybench} is long-horizon in the sense of extended dialogue histories and cross-application context, yet its reported execution traces typically involve a modest number of action steps. Mind the Gap~\cite{mindthegap} contains dense Office-operation tasks with many machine-gradable criteria, but it is derived from standardized proficiency exams rather than open-ended workplace deliverables. DocOps~\cite{docops} includes workflow-level and cross-document operations, but its reported agent runtimes are on the order of minutes rather than hours. Other benchmarks are even more clearly bounded in scope: SheetCopilot~\cite{sheetcopilot} focuses on spreadsheet-control tasks requiring a small number of atomic actions; SpreadsheetBench V1~\cite{spreadsheetbench} centers on single-spreadsheet manipulation and formula problems, although SpreadsheetBench 2~\cite{spreadsheetbench2} moves toward end-to-end business spreadsheet workflows; PPTC~\cite{pptc} evaluates PowerPoint task completion through API-call sequences; PPT-Eval~\cite{ppteval} focuses on bounded GUI-based slide-editing tasks; and SlidesBench~\cite{slidesbench} targets single-slide creation. In contrast, OmegaUse-OfficeVal targets long-horizon office-suite workflows that take roughly two hours of human labor on average.

A second limitation of existing office-oriented benchmarks is the absence of task-level economic grounding. OfficeBench~\cite{officebench}, OdysseyBench~\cite{odysseybench}, SpreadsheetBench~\cite{spreadsheetbench}, PPTC~\cite{pptc}, PPT-Eval~\cite{ppteval}, Mind the Gap/OfficeEval~\cite{mindthegap}, DocOps~\cite{docops}, and SheetCopilot~\cite{sheetcopilot} evaluate important aspects of office automation, such as multi-application workflows, spreadsheet manipulation, PowerPoint editing, document operations, and fine-grained Office proficiency. However, based on publicly available documentation, they generally do not provide per-task economic annotations comparable to those in OmegaUse-OfficeVal, such as recorded human labor time or task-price proxies. Consequently, while these benchmarks can assess whether an agent completes an office-related task, they provide limited support for analyzing performance in terms of the human effort or market value associated with each individual task.

\subsection{Computer-Use and GUI-Agent Benchmarks}
A parallel line of work evaluates GUI agents and general computer-use agents (CUAs) in interactive, execution-based environments. OmegaUse-OfficeVal differs from these benchmarks in two main respects: it is not tied to a particular interactive execution environment, and it focuses on long-horizon office-suite workflows rather than short-horizon interactions.

Early efforts focused on web-based environments, including simplified benchmarks such as MiniWoB and MiniWoB++~\cite{miniwob,miniwobpp}, the shopping-oriented WebShop~\cite{webshop}, more realistic web environments such as WebArena and VisualWebArena~\cite{webarena,visualwebarena}, and benchmarks grounded in real websites such as Mind2Web and WebVoyager~\cite{mind2web,webvoyager}. This line of work has since expanded to operating-system and desktop environments. OSWorld~\cite{osworld} introduced a scalable execution-based benchmark for open-ended computer tasks, while Windows Agent Arena~\cite{windowsarena} adapts this setting to Windows 11 virtual machines and MacOSWorld~\cite{macosworld} targets macOS. Process- and diagnostic-oriented desktop benchmarks such as OSUniverse and WindowsWorld~\cite{osuniverse,windowsworld} further study execution complexity and introduce checkpoint-based scoring that allows multiple valid execution paths. Broader agentic benchmarks such as GAIA and $\tau$-bench~\cite{gaia,taubench} also evaluate multi-step, multi-tool reasoning beyond pure GUI control. Most existing computer-use and GUI-agent benchmarks are still short-horizon: their tasks are typically designed as bounded interactions that can be completed within a small number of steps or a few minutes of human effort. Such settings are useful for evaluating perception, grounding, local UI control, and short-term tool use, but they provide limited coverage of sustained workflows that require long-term state tracking, planning, and error recovery.

OmegaUse-OfficeVal differs from these benchmarks in two main respects. First, most computer-use benchmarks are designed primarily around interactive task execution in controlled web, desktop, or operating-system environments, where GUI operation and environment-state transitions are the central objects of evaluation. In contrast, OmegaUse-OfficeVal does not prescribe or privilege a particular execution trajectory: agents may complete tasks through GUI interactions, scripts, APIs, or hybrid strategies. Evaluation instead targets the final office deliverable, including its task completion, usability, and editability. Second, unlike short-horizon GUI benchmarks, OmegaUse-OfficeVal focuses on long-term office-suite workflows whose completion requires more than two hours of human labor on average.

A notable recent exception is OSWorld 2.0~\cite{osworld2}, which reframes computer-use evaluation around long-horizon workflows. OSWorld 2.0 comprises 108 tasks whose median task takes a skilled human roughly 1.6 hours of active operation and requires hundreds of agent steps, with completion graded using many weighted checkpoints. Nevertheless, OSWorld 2.0 and OmegaUse-OfficeVal differ fundamentally in their evaluation target. OSWorld 2.0 remains a computer-use benchmark: even in the long-horizon regime, it centers on GUI operation within a controlled operating-system environment, and its evaluation is primarily process- and state-oriented. In other words, it asks whether the agent performs the right operations and reaches the expected environment state. OmegaUse-OfficeVal instead evaluates the final office deliverable itself and is agnostic to the means of production: agents are judged by the usability, editability, formatting, structure, and task correctness of the files they produce, rather than by the trajectory they follow. The two benchmarks also differ in economic grounding. OSWorld 2.0 reports human-annotated task-completion times and provides an aggregate economic-coverage analysis based on occupation-level mappings; however, this economic mapping characterizes the benchmark as a whole rather than assigning each deliverable a task-level price or market-value proxy. OmegaUse-OfficeVal instead pairs long-horizon office-suite tasks with explicit task-level economic annotations, including recorded human labor time and task-price proxies, enabling value-aware analysis of agent performance for each individual task.

\section{Task Taxonomy by Operation Intent and Domain}
\label{sec:task_taxonomy}

We present the task taxonomy with respect to operation intent in Table \ref{tab:operation_intent_taxonomy} and domain in Table \ref{tab:domain_taxonomy}. Each category is accompanied by representative examples.

\begin{table}[t]
\small
\centering
\caption{Operation-intent taxonomy with examples.}
\label{tab:operation_intent_taxonomy}
\begin{tabular}{lp{0.62\columnwidth}}
\toprule
\textbf{Label} & \textbf{Representative examples} \\
\midrule
Reformat &
``Fix the abstract format,'' ``Set the body text to 1.5 line spacing'' \\
Restructure &
``Convert the layout to A3 landscape,'' ``Reformat four pages into two pages,'' ``Compress three interface screenshots onto one A4 page'' \\
Annotate &
``Add pinyin above the text,'' ``Insert a handwritten signature'' \\
Extract &
``Organize the choir members and competition songs into Excel,'' ``Convert the data in the image into a table'' \\
Compute &
``Subtract expenses from the balance to obtain the remaining balance,'' ``Calculate payable amount by subtracting deductions and adding commission'' \\
Beautify &
``Lay out each page,'' ``Organize the content and keep it concise'' \\
Other &
Other operation intents \\
\bottomrule
\end{tabular}
\end{table}

\begin{table}[t]
\small
\centering
\caption{Domain taxonomy with examples.}
\label{tab:domain_taxonomy}
\begin{tabular}{lp{0.62\columnwidth}}
\toprule
\textbf{Label} & \textbf{Representative examples} \\
\midrule
Academic Papers &
Paper formatting, English abstracts, reference lists \\
Education \& Examination &
``Midterm teaching-quality assessment,'' answer sheets, lesson plans \\
Financial Data &
``Corporate revenue growth rate,'' attendance records, payroll calculation \\
Engineering \& Technology &
Technical roadmap diagrams, project-funding applications \\
Administrative Affairs &
Choir member lists, singing-competition finalist lists \\
Business Operations &
ROI analysis, number of advertising shoots \\
Other &
Other application domains \\
\bottomrule
\end{tabular}
\end{table}

\section{Incentive Mechanism and Computation Method for Human Labor Time}\label{apx:compute_human_labor_time}

\begin{algorithm}[t]
\footnotesize
\caption{Human Labor Time Measurement}
\label{alg:human_time_incentive}
\begin{algorithmic}[1]
\Statex \hspace*{-\algorithmicindent}\textbf{Input:} Task $t$, annotator pool $\mathcal{A}$
\Statex \hspace*{-\algorithmicindent}\textbf{Output:} Human labor time $H_t$ and task-level annotator scores
\Statex \hspace*{-\algorithmicindent}\parbox[t]{\dimexpr\linewidth+\algorithmicindent\relax}{
\textbf{Incentive policy:} To encourage efficient completion under quality constraints, task scores are aggregated daily. Among quality-valid annotators, the top $25\%$ receive $1.5\times$ the base daily salary, and those ranked between the top $25\%$ and $50\%$ receive $1.25\times$ the base daily salary.}

\State Randomly assign $t$ to two annotators $a_1,a_2 \in \mathcal{A}$
\For{each $a_i \in \{a_1,a_2\}$}
    \State Record completion time $\tau_i$
    \State A senior expert to check whether the submitted artifact satisfies requirements
    \If{the artifact fails the expert-assessed quality check}
        \Repeat
            \State Request revision and include revision time in $\tau_i$
            \State Recheck whether the revised artifact satisfies quality control
        \Until{the artifact satisfies quality control}
        \State Mark the submission valid and set $s_i \gets 0$
        \State Mark $a_i$ as revised
    \Else
        \State Mark the submission valid
        \State Mark $a_i$ as unrevised
    \EndIf
\EndFor

\If{both $a_1,a_2$ have valid submissions}
    \If{exactly one annotator revised}
        \State Set the revised annotator's score to $0$ and the unrevised one's score to $1.5$
    \ElsIf{both annotators revised}
        \State Set both annotator scores to $0$
    \ElsIf{$\max(\tau_1,\tau_2) > 1.3 \cdot \min(\tau_1,\tau_2)$}
        \State Assign score $2$ to the faster annotator and score $1$ to the slower annotator
    \Else
        \State Assign score $1.5$ to both annotators
    \EndIf

    \If{$\max(\tau_1,\tau_2) > 1.3 \cdot \min(\tau_1,\tau_2)$}
        \State Assign $t$ to a third annotator $a_3$
        \State Record $\tau_3$ and apply the same quality-control procedure
        \If{$a_3$ revised}
            \State Assign score $0$ to $a_3$
        \ElsIf{$\tau_3 < 0.7 \cdot \min(\tau_1,\tau_2)$}
            \State Assign score $2$ to $a_3$
        \ElsIf{$\tau_3 > 1.3 \cdot \max(\tau_1,\tau_2)$}
            \State Assign score $1$ to $a_3$
        \Else
            \State Assign score $1.5$ to $a_3$
        \EndIf
    \EndIf
\EndIf

\State Let $\mathcal{T}_{\mathrm{valid}}$ be all valid completion times for task $t$
\State Compute $H_t$ as the average of the two shortest values in $\mathcal{T}_{\mathrm{valid}}$

\State \Return $H_t$
\end{algorithmic}
\end{algorithm}

Here we introduce the details about the quality-gated incentive mechanism and computation method for task-level human labor time. The full procedure of the incentive mechanism and computation method is described in Algorithm \ref{alg:human_time_incentive}.

First, we define an incentive policy that assigns task-level scores to annotators. Annotator scores are aggregated daily to determine bonus payments. Among annotators whose deliverables satisfy the quality requirements, those ranked in the top 25\% on a given day receive 1.5 times the base daily salary, while those ranked between the top 25\% and 50\% receive 1.25 times the base daily salary. This incentive mechanism encourages efficient task completion, while the quality gate and revision process discourage low-quality or rushed submissions.

Second, each task is randomly assigned to two annotators, and their completion times are recorded separately (Lines 1--14). If a submitted deliverable fails the expert quality assessment, the annotator is asked to revise it, and the revision time is included in the total completion time for that task. Once the revised deliverable passes the quality assessment, the submission is treated as valid for time measurement, but the annotator receives a task score of 0 because the initial submission failed quality control. We note that this quality assessment involves some degree of expert judgment, since it is impractical to manually verify every detail of each artifact during this stage.

Third, we assign scores to annotators based on revision status and relative completion time (Lines 15--23). When both annotators submit valid deliverables for the same task, we first check whether either annotator required revision. If exactly one annotator revised their submission, the revised annotator receives a score of 0 and the unrevised annotator receives a score of 1.5. If both annotators required revision, both receive a score of 0. If neither annotator required revision, we compare their completion times. If the slower time is more than 1.3 times the faster time, the faster annotator receives a score of 2 and the slower annotator receives a score of 1; otherwise, both annotators receive a score of 1.5.

When the two valid completion times differ substantially, i.e., when the slower time is more than 1.3 times the faster time, we recruit a third annotator to complete the same task and obtain an additional timing measurement (Lines 24--34). The third annotator follows the same quality-control procedure. If the third annotator requires revision, the third annotator receives a score of 0. Otherwise, the third annotator's completion time is compared with the two original valid completion times: the third annotator receives a score of 2 if their time is less than 70\% of the shorter original time, a score of 1 if their time is greater than 1.3 times the longer original time, and a score of 1.5 otherwise.

\section{Code-Based Verification Protocol Details}
\label{app:code_based_verification_details}

We build the verification protocol through an iterative process. First, we generate task-specific rubrics by combining LLM assistance with
expert revision. Next, we use a coding agent to translate the rubrics into executable verification code. Finally, we conduct human-code discrepancy resolution to revise both the rubrics and the verifiers. We iterate this process until the code-based scores are sufficiently aligned with expert judgment.

\subsection{Rubric Generation with Expert Revision}

We construct fine-grained scoring rubrics and refine them through expert revision. We first use an LLM to generate an initial rubric based on the task objective, input files, and expected deliverable. We then conduct in-depth expert revision to iteratively improve the correctness, coverage, and evaluability of the rubric. Each task rubric consists of two main dimensions: a usability rubric and a task completion rubric.

The \textbf{usability rubric} determines whether the delivered artifact satisfies basic usability requirements. If the artifact fails to meet these requirements, the task receives a score of zero, and the task completion rubric is not further evaluated. For example, we check whether the file format is correct, whether the file can be opened normally, whether the content or layout is severely corrupted, and whether the file remains editable and modifiable. This dimension prevents models from receiving credit for files that appear complete but cannot actually be used or further modified by the user.

When designing \textbf{task completion rubric} items, OmegaUse-OfficeVal follows a user-centered scoring principle: rubrics reward completed task-specific requirements while penalizing unintended changes that increase the user's repair burden. For each task, positive scoring items are derived from the explicit requirements and assign credit when a requested modification, transformation, or output is correctly completed. In addition, we specify task-relevant negative scoring items for likely or observed failure modes, such as disrupted layouts, altered content that should have been preserved, inconsistent formatting, missing sections, rasterized editable content, or other changes that increase the user's correction cost. Thus, we evaluate the final artifact from the perspective of a user who receives the file, measuring both how much useful work the agent has completed and whether the delivered artifact remains practically usable with minimal additional human repair.

Each rubric item is assigned a discrete weight according to the estimated time and difficulty required for a human to complete or repair it. For positive scoring items corresponding to required task components, we use a three-level scale: simple items receive $+1$ point, medium-difficulty items receive $+3$ points, and difficult items receive $+5$ points. For negative scoring items corresponding to unintended changes or avoidable damage to content that should have been preserved, we use the symmetric penalty scale of $-1$, $-3$, and $-5$, based on the estimated difficulty of repairing the issue.

During expert revision, experts remove vague or unverifiable evaluation items, add missing task-critical requirements, balance the weights of different scoring points, and ensure that the rubric reflects genuine task completion quality rather than superficial features alone.

\subsection{Verification Code Generation and Reflection.}

After experts complete rubric revision, we convert the scoring rubrics into verification code. Depending on the task type and output file format, the code verifier automatically checks whether the final delivered artifact satisfies the rubric and reports the corresponding score.
The code verifier is generated with the assistance of coding agents and subsequently revised by experts. Specifically, a coding agent first generates initial verification code based on the expert-defined rubric described in the previous section. We then provide both the rubric and the generated verifier to the coding agent again, prompting the LLM to reflect on potential mismatches between the rubric and the verification code. Based on this reflection output, experts inspect the code logic item by item, checking whether each scoring rule accurately corresponds to the rubric and whether there are omissions, false positives or negatives, or implementation deviations. For subjective items that cannot be reliably verified by code, experts further revise the rubric to express them as more concrete, observable, and executable properties of the deliverable. After the reflection and expert revision, we generate the verification code again by the coding agent. This process ensures that tasks included in OmegaUse-OfficeVal are not only grounded in realistic user requests, but also support stable and reproducible code-based automated evaluation.

\subsection{Human-Code Discrepancy Resolution}

To ensure the quality of the code verifiers, we further conduct a human-code discrepancy resolution process after LLM reflection and expert inspection of the code logic. Specifically, for each task, experts manually score agent-generated artifacts and compare their scores with the scores produced by the code-based verifier. Experts first inspect the final deliverable generated by an LLM agent according to the scoring rubric, determine which rubric items are satisfied, and assign the corresponding score. We then run the code verifier on the same artifacts to obtain the automated grading results.

We compare expert scores and verifier scores at the level of individual rubric items and analyze the sources of discrepancy. If a discrepancy is caused by an implementation error, we revise the verifier logic. If it arises from an ambiguous rubric description, we return to the rubric revision stage. If it reflects inconsistent expert interpretation, we further calibrate the scoring standard. If it is due to the existence of multiple reasonable ways to complete the task, we extend the verifier logic to cover different but valid approaches.

Through this cycle of human-code discrepancy analysis, rubric correction, and verifier revision, we progressively improve the consistency between automated evaluation and expert judgment. This process ensures that code verifier-based evaluation results are reproducible while remaining closely aligned with human expert assessments grounded in the original user requirements.

\section{More Experiment Settings}\label{apx:exp_setting}

\subsection{Agent Scaffold.}
All models are evaluated using the same in-house agent scaffold. The scaffold supports programmatic tool use, shell execution, and file operations, enabling agents to inspect, create, and modify office artifacts through scripts and file-level APIs. GUI-level computer-use capabilities are not included in the current evaluation. This setup focuses on whether models can complete office-suite tasks through programmatic file interaction, rather than through visual control of office applications.

\subsection{Execution Environment.}
All LLM evaluations are conducted in Docker containers running Ubuntu 24.04.1 LTS with Python 3.10. The environment is CPU-only, and model inference is performed through remote model APIs. Each evaluation container is allocated up to 108 CPU-core equivalents and 300 GiB of memory. It also includes LibreOffice 24.2 for office document processing and rendering. The token cost is calculated according to the token pricing of the corresponding LLM. Because agent runtime can be affected by the scaffold implementation, execution environment, system load, and remote API latency, the reported LLM time cost should be interpreted as a reference measurement rather than as an exact estimate.

\subsection{Inference Configuration.}
For each task, the agent receives the task instruction and associated input files, and is required to produce the final deliverable artifacts in the specified output format. We use the same system prompt and task-prompt template for all models as shown in Appendix \ref{app:agent_task_prompt}. Each task is run with a wall-clock timeout of 14,400 seconds. We execute 10 task environments in parallel, with at most two concurrent inference threads per model. All artifacts are evaluated using the code-based verifiers described in Section~\ref{sec:code-based-verification}.

\subsection{Prompt for Task Evaluation}
\label{app:agent_task_prompt}

For each task, the agent is provided with a standardized task prompt that specifies the target file types, the user instruction, the required output location, and constraints on using LibreOffice within the sandboxed execution environment. The prompt is designed to preserve the open-ended nature of the workflow while ensuring that final deliverables remain directly accessible for evaluation. The template is shown below.

\begin{tcolorbox}[
    colback=gray!5,
    colframe=gray!60,
    boxrule=0.5pt,
    arc=1pt,
    left=4pt,
    right=4pt,
    top=4pt,
    bottom=4pt,
    breakable
]
\small
\ttfamily
\setlength{\parindent}{0pt}
Please modify the \{FILE\_TYPES\} files in the current directory according to the following requirements: \{TASK\_INSTRUCTION\}.

\vspace{0.5em}

All final deliverable files must retain their original file formats. Keep every final deliverable file directly in the current directory; do not place deliverables in subdirectories.

\vspace{0.6em}

LibreOffice is available through managed `libreoffice' and `soffice' commands. Invoke those commands by name as top-level Bash commands when needed; do not launch them through Python `subprocess', a nested `bash'/`sh' command, or an absolute executable path, because only the top-level managed command receives the sandbox exemption. Each task already has an isolated LibreOffice profile. Never use `pkill', `killall', or another process-wide cleanup command for LibreOffice; wrap only the individual LibreOffice invocation with `timeout' if a conversion may hang.
\end{tcolorbox}

\section{Evaluation Metric Definitions}
\label{apx:evaluation_metrics}
In the experimental evaluation, we report three main metrics. The first is the overall score, which measures overall completion quality across the benchmark. For each task $t$, we use the task score $\mathrm{Score}(t)$ defined in Section~\ref{sec:code-based-verification}. Given the set of benchmark tasks $\mathcal{T}$, the overall score is
\[
\mathrm{Score}
=
\sum_{t \in \mathcal{T}} \mathrm{Score}(t).
\]

To evaluate performance under different notions of economic value, we additionally report two value-weighted metrics. The \textbf{time-weighted score} measures the amount of task completion achieved after weighting each task by its associated human labor time. Let $H_t$ denote the recorded human labor time for task $t$. The time-weighted score is defined as
\[
\mathrm{Score}_{\mathrm{time}}
=
\sum_{t \in \mathcal{T}} H_t \cdot \mathrm{Score}(t).
\]
This metric gives greater influence to tasks that require more human effort and can be interpreted as the total human labor time covered by the agent's partial task completion.
The \textbf{price-weighted score} measures the amount of task completion achieved after weighting each task by its task price proxy. Let $P_t$ denote the task price proxy for task $t$. The price-weighted score is defined as
\[
\mathrm{Score}_{\mathrm{price}}
=
\sum_{t \in \mathcal{T}} P_t \cdot \mathrm{Score}(t).
\]
This metric gives greater influence to tasks with higher estimated market value. Together, these metrics allow us to compare models not only by average quality, but also by the amount of human labor time or market value they capture across the benchmark.

\begin{figure}[t]
    \centering
    \includegraphics[width=0.8\columnwidth]{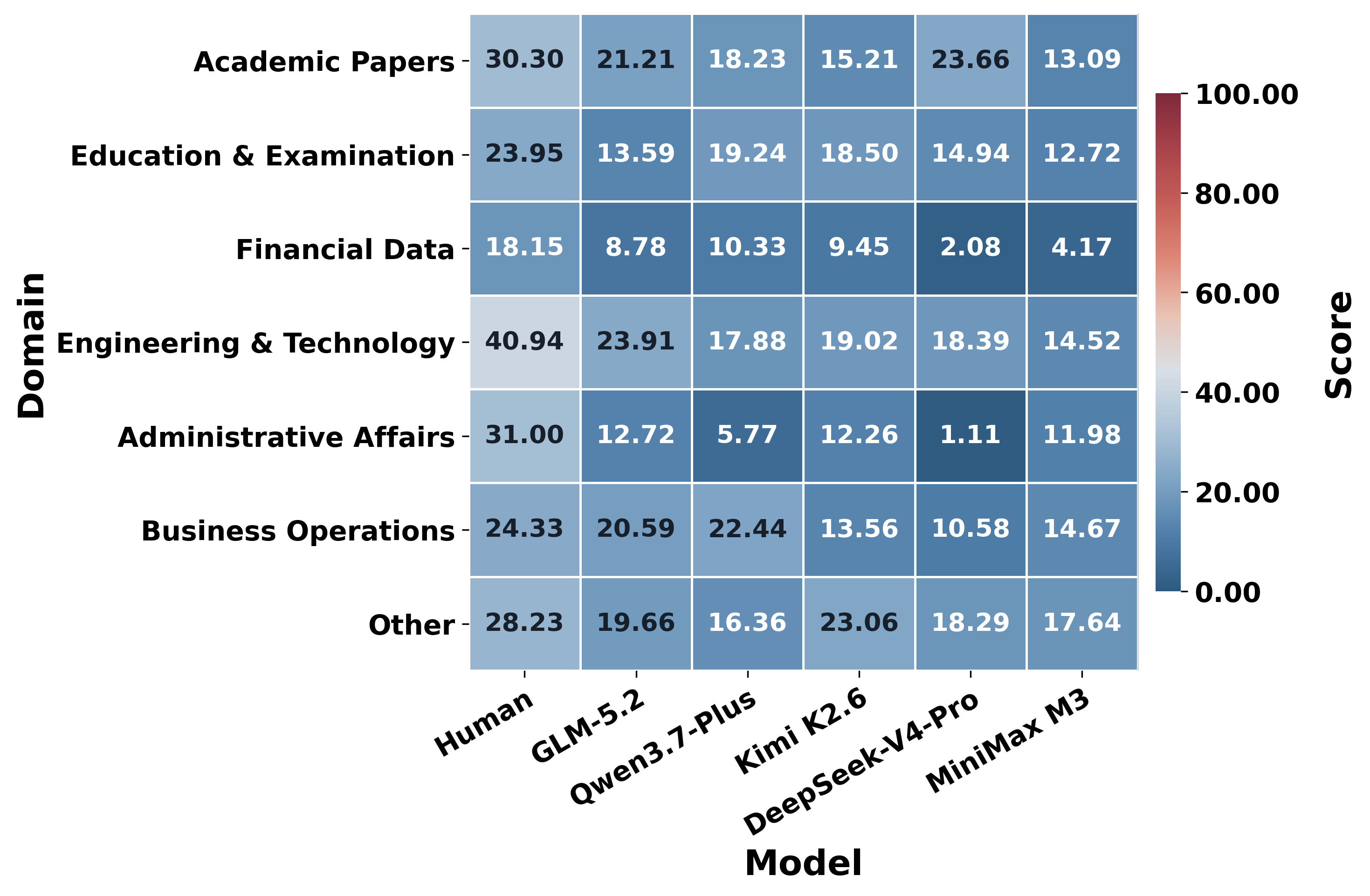}
    \caption{Task scores by domain. Each cell reports the average score for tasks in the domain.}
    \label{fig:domain_score_heatmap}
\end{figure}

\begin{figure}[t]
    \centering
    \includegraphics[width=0.8\columnwidth]{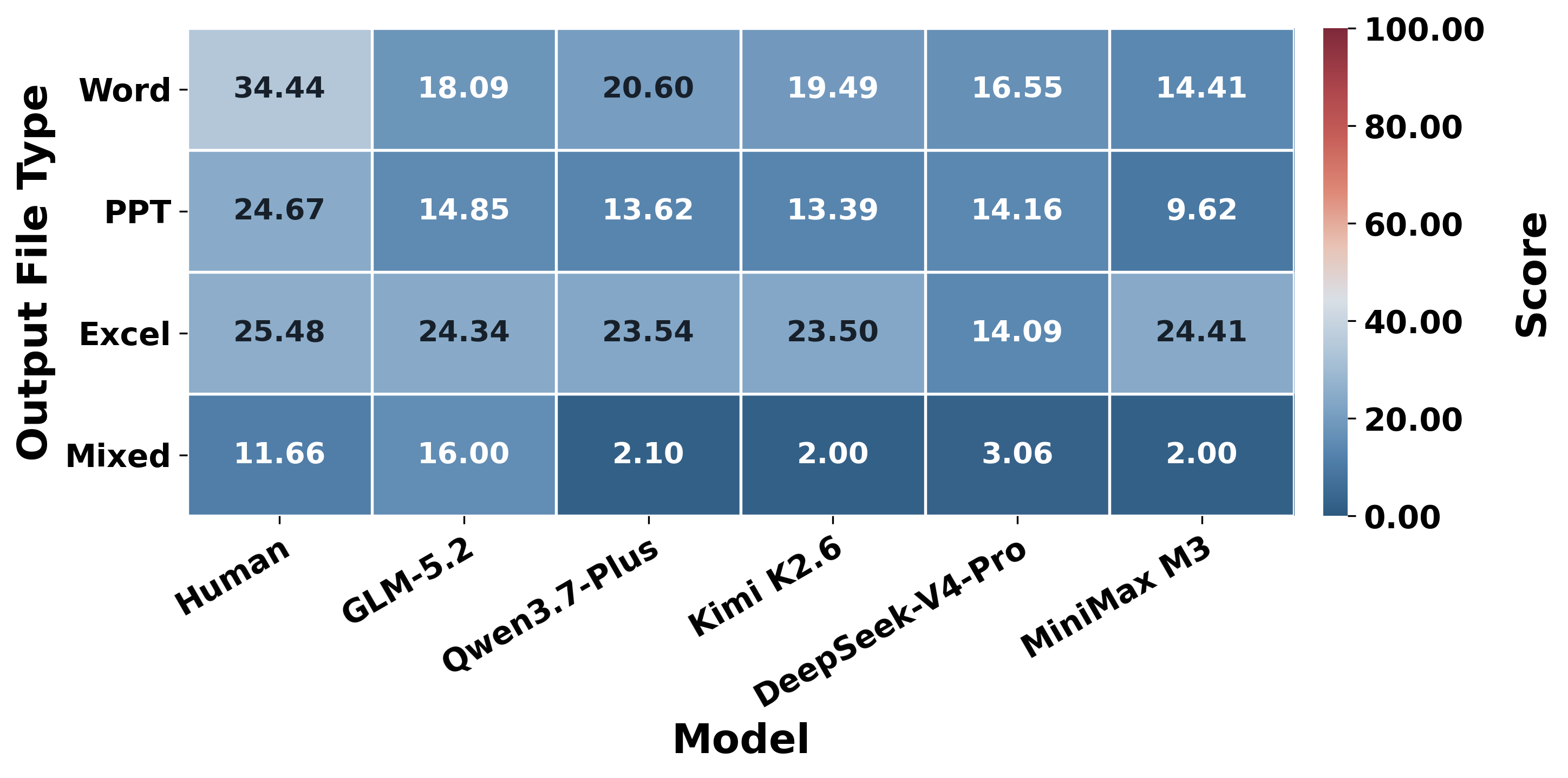}
    \caption{Task scores by output file type. Each cell reports the average score for tasks with the output file type.}
    \label{fig:filetype_score_heatmap}
\end{figure}

\begin{figure}[t]
    \centering
    \includegraphics[width=0.8\columnwidth]{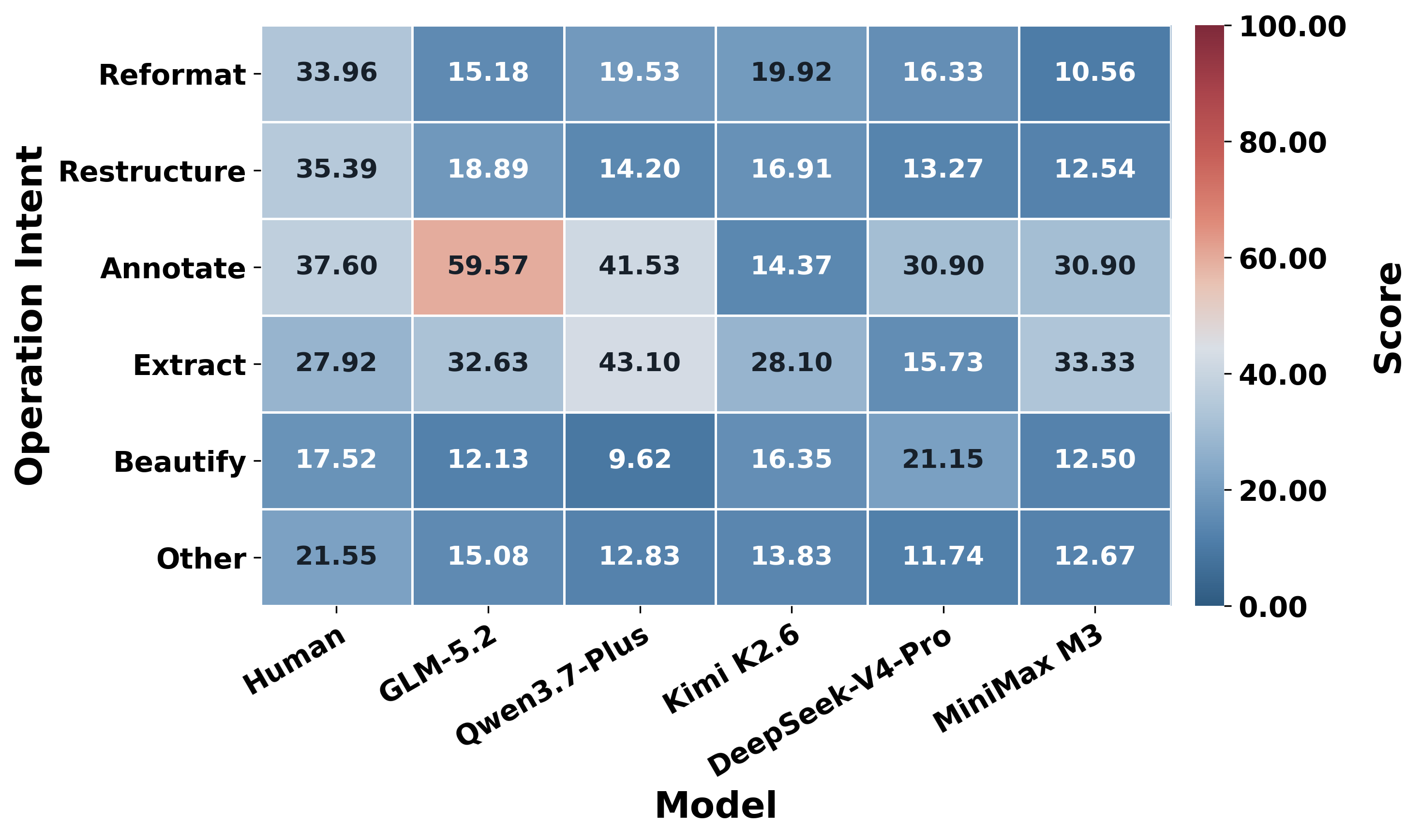}
    \caption{Task scores by operation intent. Each cell reports the average score for tasks with the operation intent.}
    \label{fig:operation_intent_score_heatmap}
\end{figure}

\section{Additional Breakdown Results of Performance}
\label{app:additional_results}

Figure~\ref{fig:domain_score_heatmap} shows model performance across different domains. Human annotators achieve the best performance in most domains, confirming that the benchmark remains challenging for current LLMs across a broad range of office-suite scenarios. Among the LLMs, however, the best-performing model varies by domain. GLM-5.2 performs best on Engineering \& Technology tasks, while Qwen3.7-Plus achieves the best LLM score on Business Operations. DeepSeek-V4-Pro performs best on Academic Papers, and Kimi K2.6 achieves the strongest result in the Other category, suggesting relatively better robustness on long-tail task types. MiniMax M3 performs comparatively better on Administrative Affairs than on several other domains. Financial Data is particularly challenging for most LLMs. These domain-level differences suggest that current LLMs do not exhibit uniform capabilities across office-suite workflows.

Figure~\ref{fig:filetype_score_heatmap} and Figure \ref{fig:operation_intent_score_heatmap} provide additional breakdowns of model scores by output file type and operation intent. These results show that model performance varies substantially across task characteristics. For output file type, models generally perform better on Word and Excel tasks than on PPT and mixed-file tasks. Excel tasks are relatively more tractable for several models, with performance approaching that of human annotators in some cases, likely because many spreadsheet operations can be expressed and executed through programmatic file manipulation.
For operation intent, annotation and extraction tasks show the highest scores for several models. GLM-5.2 performs particularly well on annotation tasks, while Qwen3.7-Plus performs strongly on extraction tasks. In contrast, beautification and other loosely specified tasks tend to be more difficult, likely because they require layout judgment, visual consistency, and preservation of existing content.

\end{CJK}

\end{document}